\documentclass[10pt,twocolumn,letterpaper]{article}

\usepackage{wacv}              

\usepackage{graphicx}
\usepackage{amsmath}
\usepackage{amssymb}
\usepackage{booktabs}
\usepackage{multirow}
\usepackage{float}
\usepackage[labelformat=simple, labelfont=Huge, labelsep=period]{subcaption}

\usepackage[pagebackref,breaklinks,colorlinks]{hyperref}

\usepackage[capitalize]{cleveref}
\crefname{section}{Sec.}{Secs.}
\Crefname{section}{Section}{Sections}
\Crefname{table}{Table}{Tables}
\crefname{table}{Tab.}{Tabs.}

\def\wacvPaperID{*****} 
\def\confName{WACV}
\def\confYear{2025}

\begin{document}

\title{LS-GAN: Human Motion Synthesis with Latent-space GANs}

\author{Avinash \\ Amballa \and
Gayathri \\ Akkinapalli \\
University of Massachusetts Amherst\\
{\tt\small \{aamballa,gakkinapalli,vmuralikrish\}@umass.edu}
\and Vinitra \\ Muralikrishnan  
}



\maketitle

\begin{figure*}[h]
\centering
\resizebox{\linewidth}{!}{%
    \begin{tabular}{cccc}
        \subfloat[\LARGE a man kicks with something or someone with his left leg. (Vanilla GAN ) ]{{\includegraphics[width=10cm,trim=300 300 300 50,clip]{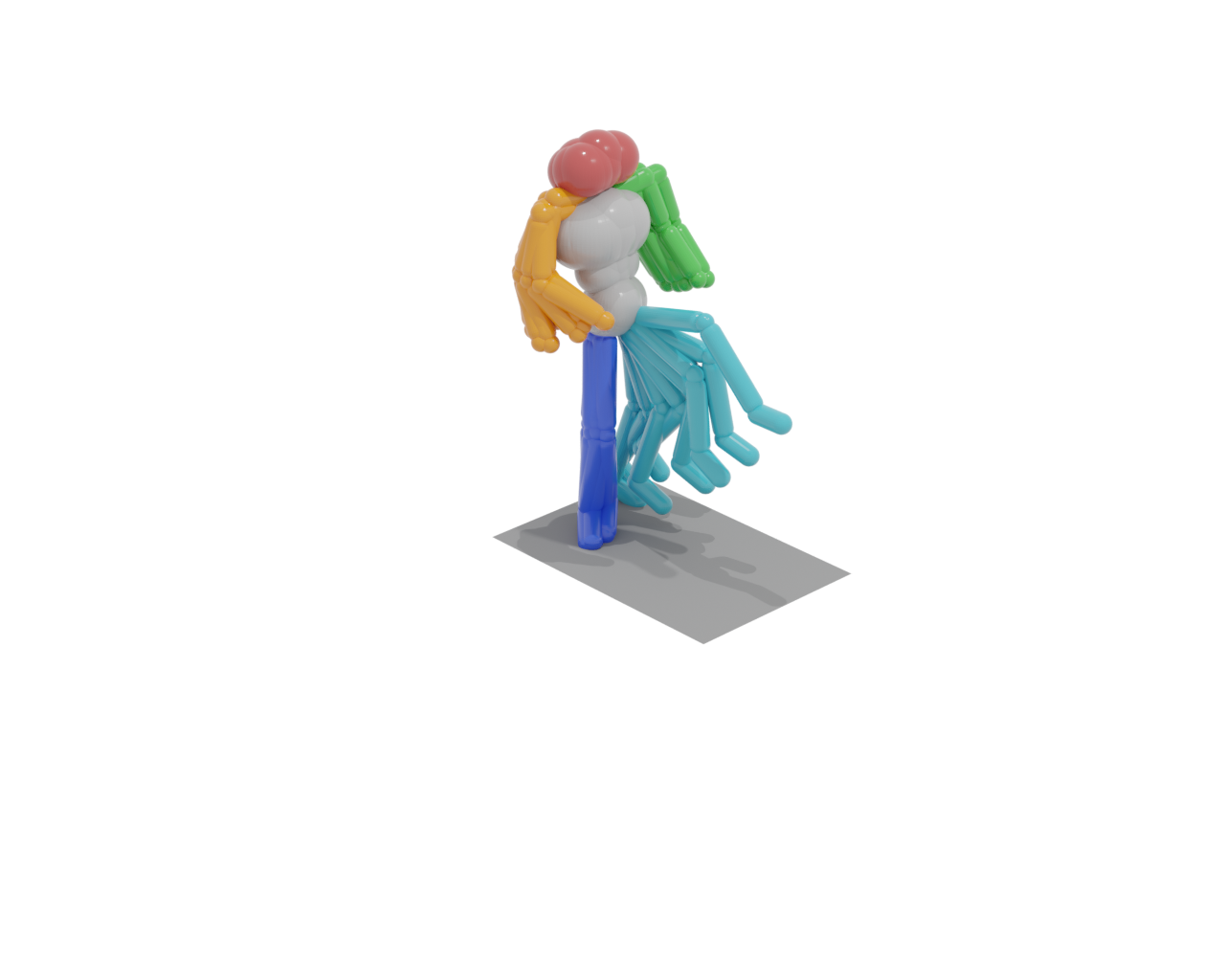} }} &
         \subfloat[\LARGE
 a person raised arms up and pull them down. (Deep GAN)]{{\includegraphics[width=10cm,trim=300 300 300 50,clip]{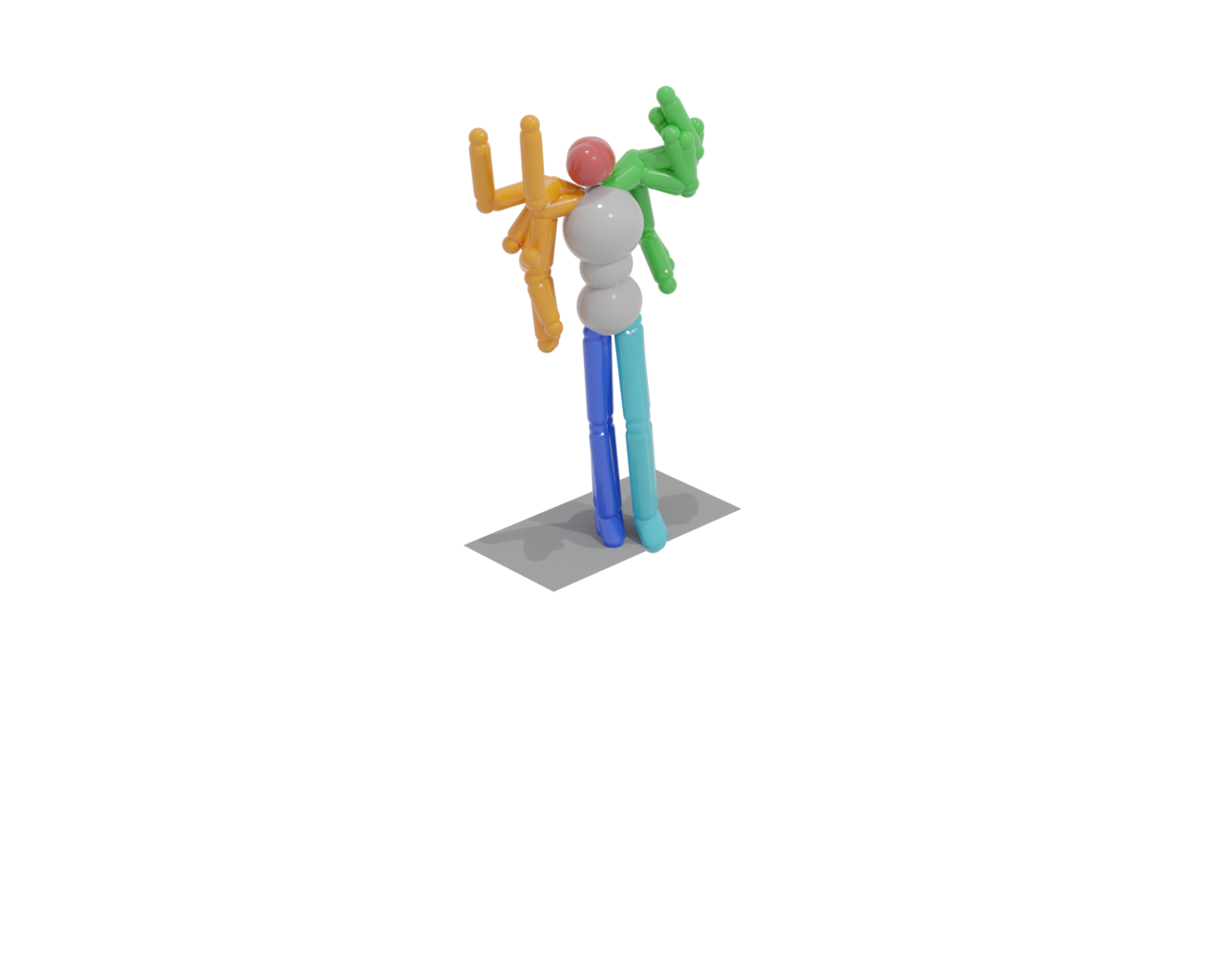} }} &
         \subfloat[\LARGE
 a person lifts up their left arm to the side. (Vanilla WGAN-GP)]{{\includegraphics[width=10cm,trim=300 300 300 50,clip]{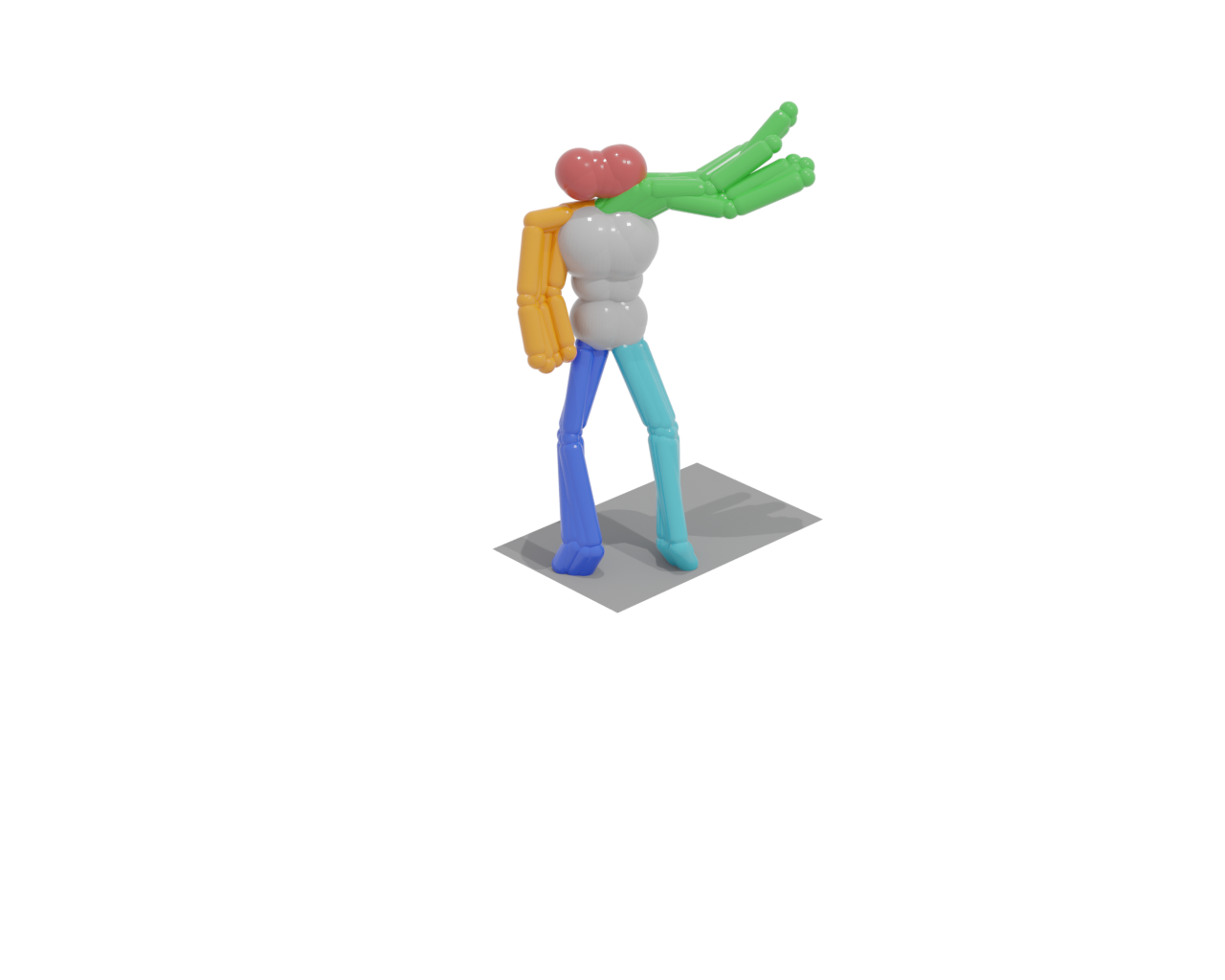} }} &
         \subfloat[ \LARGE
 a person walks in a circle to their right.  (Deep WGAN-GP ) ]{{\includegraphics[width=10cm,trim=150 200 200 50,clip]{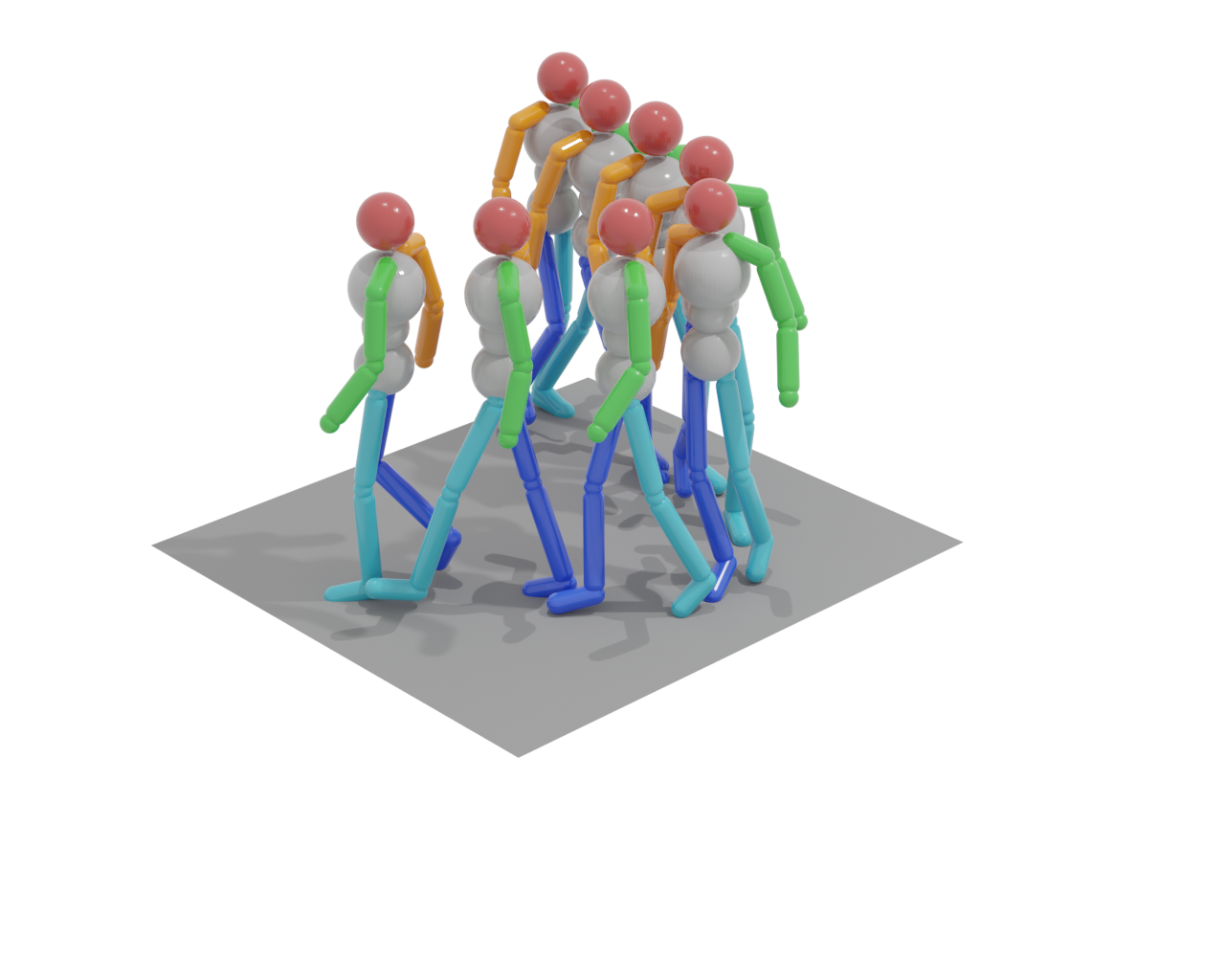} }} 
    \end{tabular}}
    \caption{Qualitative results of text-to-motion shown by LS-GAN}
    \label{fig:1}
\end{figure*}

\begin{figure*}[!ht]
\centering
\resizebox{\textwidth}{!}{%
    \begin{tabular}{ccc}
        \subfloat[\LARGE a person drinks water.  ]{{\includegraphics[width=10cm, trim=200 300 300 50,clip]{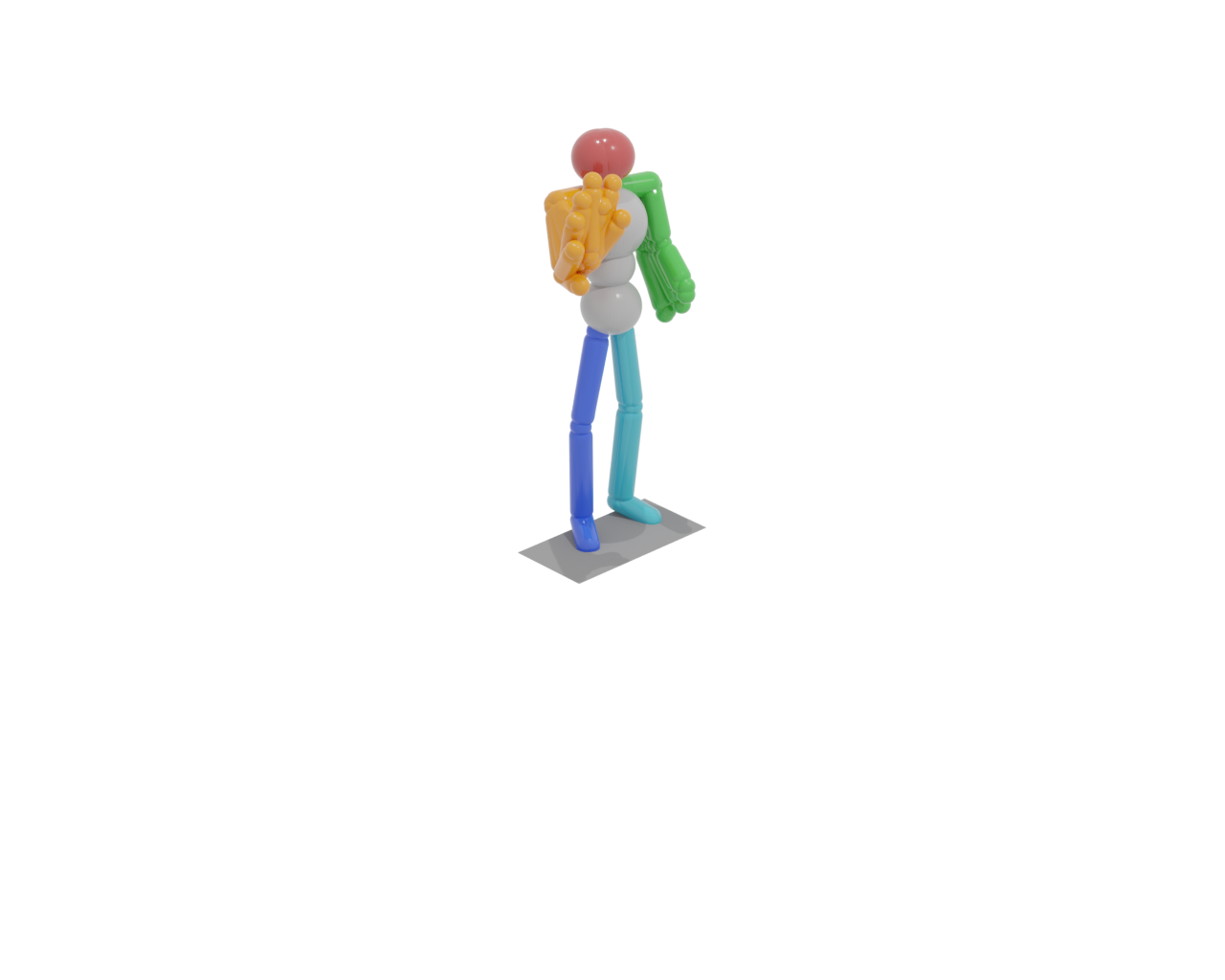} }} 
        \subfloat[\LARGE a person doing jumping jacks.   ]{{\includegraphics[width=10cm, trim=200 300 200 50,clip]{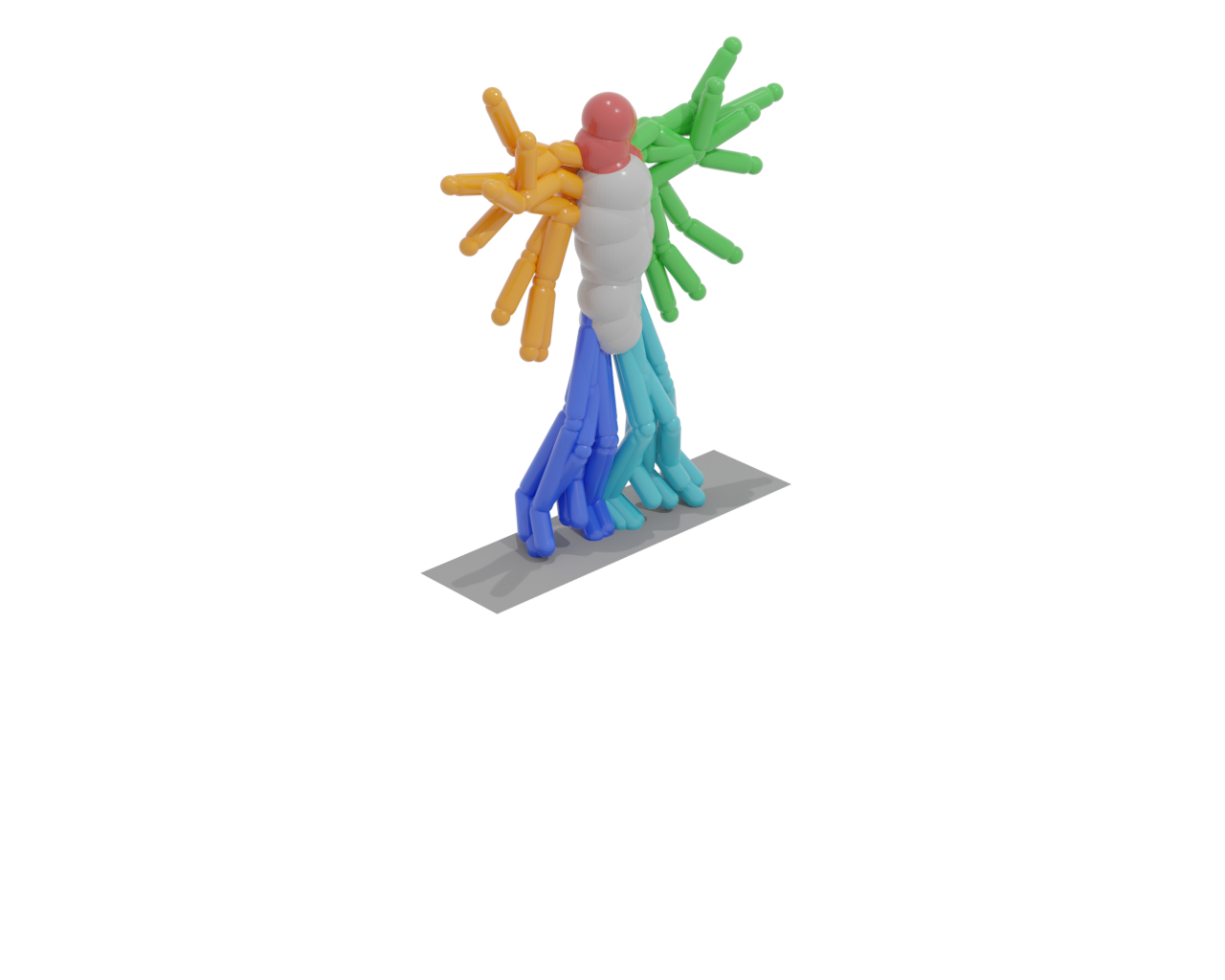} }} 
        \subfloat[\LARGE a person jogs straight forward. ]{{\includegraphics[width=10cm, trim=200 200 100 50,clip]{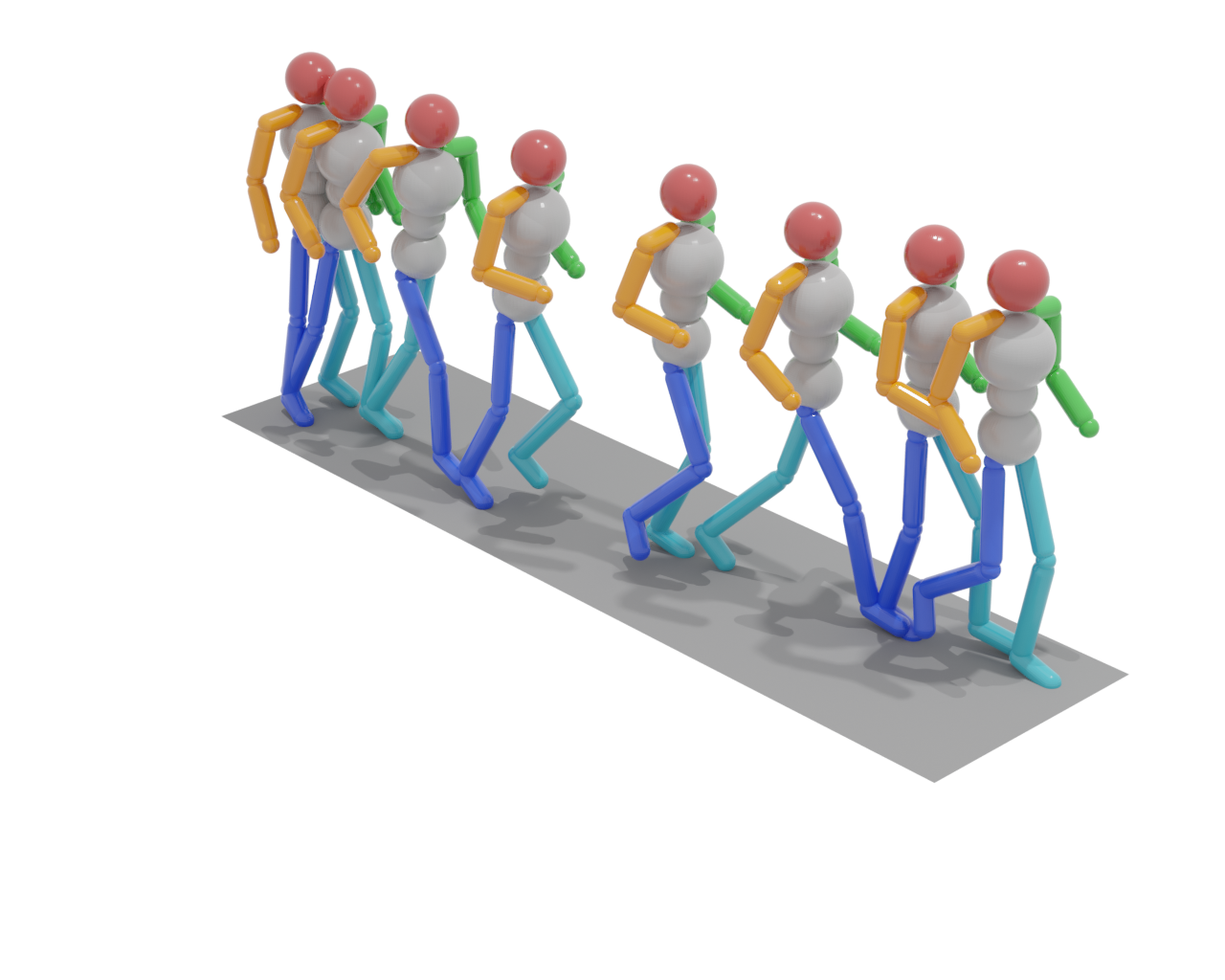} }} &
    \end{tabular}}
    \caption{Qualitative results of text-to-motion shown by our best model (Deep WGAN-GP)}
    \label{fig:7}
\end{figure*}

\begin{abstract}
    Human motion synthesis conditioned on textual input has gained significant attention in recent years due to its potential applications in various domains such as gaming, film production, and virtual reality. Conditioned Motion synthesis takes a text input and outputs a 3D motion corresponding to the text. While previous works have explored motion synthesis using raw motion data and latent space representations with diffusion models, these approaches often suffer from high training and inference times. In this paper, we introduce a novel framework that utilizes Generative Adversarial Networks (GANs) in the latent space to enable faster training and inference while achieving results comparable to those of the state-of-the-art diffusion methods. We perform experiments on the HumanML3D, HumanAct12 benchmarks and demonstrate that a remarkably simple GAN in the latent space achieves a \textbf{FID of 0.482} with more than \textbf{91\%} in FLOPs reduction compared to latent diffusion model. Our work opens up new possibilities for efficient and high-quality motion synthesis using latent space GANs.
\end{abstract}

\section{Introduction}
\label{sec:intro}

Human motion synthesis has recently seen rapid advancements in a multi-modal generative fashion, fueled by various conditional inputs such as music \cite{Li2021AICM, Li2021DanceFormerMC,Valadares2000DancingTT }, action categories \cite{Petrovich2021ActionConditioned3H, Guo2020Action2MotionCG}, and notably, natural language descriptions \cite{Guo2022GeneratingDA, Petrovich2022TEMOSGD, Tevet2022HumanMD, Guo2022TM2TSA, Ahuja2019Language2PoseNL, Kim2022FLAMEFL}. This field significantly enhances industries like gaming, film production, and virtual/augmented reality, with text-based conditioning standing out for its convenience and interpretability. However, learning a probabilistic mapping function from textural descriptors to motion sequences is challenging \cite{tevet2022motionclip} and this mapping often leads to misalignments and high computational demands due to stark differences in distributions between language descriptors and motion sequences, making the task of probabilistic mapping complex.

Conditional diffusion models \cite{zhang2022motiondiffuse, Tevet2022HumanMD, Kim2022FLAMEFL} address this problem by learning a more powerful probabilistic function from the textual descriptors to  motion sequences. However, diffusion models in raw sequential data require  computational overhead in both traning and inference. To overcome this, 
motion latent diffusion (MLD) \cite{chen2023executing} address these issues by encoding motion in a latent space using a Variational Autoencoder (VAE). However, MLD relied on computationally intensive diffusion processes to achieve high-quality image sampling, especially during the training and inference phases.

To efficiently model the motion synthesis, we propose substituting the diffusion model \cite{ho2020denoising} in the latent space with a Generative Adversarial Network (GAN) \cite{goodfellow2014generative}  to capitalize on its efficient adversarial training dynamics. Recognizing the effectiveness of GANs in learning complex representations across diverse modalities \cite{karras2019stylebased, sauer2023stylegantunlockingpowergans}, and their efficiency in training and inference compared to diffusion models, we propose to utilize them within this latent space. By leveraging GANs, we aim to accelerate the mapping between text embeddings and latent space, thus producing higher-quality motion sequences more efficiently. 


Specifically, this work undertakes the task of text-to-motion and action-to-motion synthesis using conditional Generative Adversarial Networks \cite{mirza2014conditional} in latent space, as depicted in the accompanying figure:\ref{fig:1}. We employ a Variational Autoencoder (VAE) to transition from motion space to latent space and utilize pre-trained CLIP models from MLD \cite{chen2023executing} to condition on textual input. We experiment with various GAN architectures, including vanilla GAN, deep GAN, with loss functions such as cross-entropy and Wasserstein \cite{gulrajani2017improved} to optimize performance and fidelity in generated motion sequences. Our experiment results on  HumanML3D \cite{guo2022generating} benchmark suggest that a simple GAN architecture achieves an FID of 0.482 with $91\%$ in FLOPS reduction compared to MLD. In addition, our method shows competitive performance on action-to-motion HumanAct12 \cite{Guo2020Action2MotionCG} benchmark. This strategic shift of GANs in latent space not only addresses the computational inefficiencies associated with previous diffusion-based models but also leverages the rapid generative capabilities of GANs to enhance the quality and diversity of motion synthesis, suitable for real-time applications.


\section{Related work.} 

\textbf{Motion Synthesis} is broadly categorized into conditional and unconditional motion synthesis. Unconditional motion synthesis models the entire motion space without requiring specific annotations, is discussed by Raab et al. \cite{raab2022modi} in an unsupervised setting using unstructured and unlabeled datasets. Conditional motion synthesis, on the other hand, employs inputs from various modalities such as music \cite{li2021ai} and text \cite{kim2023flame} to generate motion sequences. Text-to-motion synthesis, in particular, has become a dominant area of research due to the user-friendly nature of natural language interfaces. Additional recent advancements in the field include the development of joint-latent models like TEMOS \cite{petrovich2022temos} and conditional diffusion models \cite{zhang2022motiondiffuse, tevet2022human, kim2023flame}, which have led to significant progress. TEMOS, uses a VAE architecture to create a shared latent space for motion and text based on a Gaussian distribution. 



Motion diffuse \cite{zhang2022motiondiffuse} is the first text-based
motion diffusion model with fine-grained instructions on
body parts. MDM \cite{mdm2022human} proposes a motion
diffusion model on raw motion data to learn the relation
between motion and input conditions. 
Our work closely relates to the Motion Latent Diffusion (MLD) model \cite{chen2023executing} which utilizes a Variational Autoencoder (VAE) to encode human motion sequences into a low-dimensional latent space and decode them back to motion sequences. The MLD model then employs diffusion processes in this latent space, inspired by other latent diffusion models \cite{rombach2022highresolution}. To condition the motion sequences on specific inputs like text or actions, the model utilizes CLIP encodings \cite{radford2021learning}, demonstrating robust performance on tasks such as text-to-motion and action-to-motion. 

Moreover, approaches like MotionGPT \cite{jiang2023motiongpt} integrates language modeling for both motion and text, treating human motion as a distinct language to construct a generalized model capable of executing various motion tasks through VQ-VAE \cite{oord2018neural}. T2M-GPT \cite{zhang2023t2mgptgeneratinghumanmotion} uses a standard 1D convolutional network to map motion sequences to discrete code indices, followed by standard GPT-like model is learned to generate sequences of code indices from pre-trained text embedding. The use of GAN networks for motion synthesis has been done in Ganimator \cite{Li_2022} but uses an additional motion sequence as conditional input. On the other hand, Shiobara et al. \cite{10.1145/3458380.3458428} train Wasserstein GAN directly on the raw motion sequences. Actformer \cite{xu2022actformerganbasedtransformergeneral} proposed a GAN-based Transformer to generate motion sequence from actions. In addition, Text2Action \cite{ Ahn2017Text2ActionGA} proposed a generative model which learns the relationship between language and human action in order to generate a human action sequence.

\section{Method}

While diffusion models have shown tremendous promise and exhibit state-of-the-art performance they are expensive to train, requiring a huge corpus of data. The use of latent space in MLD \cite{chen2023executing} opens up avenues for other architectures such as GANs to also leverage it.  Specifically, given an input condition $c$ describing a motion, our Latent space GAN (LS-GAN) aims to generate a human motion $\hat{x}^{1:L}$ where L represents the motion length.




\subsection{VAE and CLIP} Our VAE architecture is borrowed from the MLD\cite{chen2023executing}, which uses transformer model as Encoder $\mathcal{E}$ and Decoder $\mathcal{D}$ with skip connections. The motion encoder $\mathcal{E}$ encodes the motion sequences, $x^{1:L}$ into a latent $z = \mathcal{E}(x^{1:L})$ , and the 
decode $z$ into the motion sequences using the decoder
$\mathcal{D}$, i.e., $\hat{x}^{1:L}$ = $\mathcal{D}(z)$ = $\mathcal{D}(\mathcal{E}(x^{1:L}))$. VAE is trained in a similar fashion as MLD with the MSE and KL divergence loss. After training, the VAE is kept fixed. We use pretrained CLIP-ViT-L-14 \cite{Radford2021LearningTV} text encoder to map text prompt. On the other hand to condition on action, we use the learnable embedding for each action category.


\subsection{Latent space GAN} 

We chose GANs for 3 reasons - (1) their effectiveness in learning complex representations across diverse modalities \cite{karras2019stylebased, sauer2023stylegantunlockingpowergans}, (2) the flexibility of implementing any architecture for the generator and discriminator and the potential adversarial training offers, (3) reduced training and inference time compared to Diffusion models. We discuss the GAN challenges in section \ref{sec:6}

\begin{figure}[t]
\centering
       {\includegraphics[width=8cm]{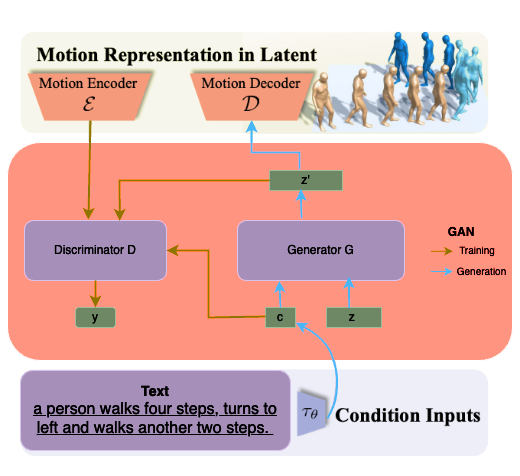} }
    \caption{ Method overview: The Generator $G$ maps the conditional input $[z,c]$ to a latent $z'$. The Discriminator $D$ learns to differentiate real $\mathcal{E}(x)$ vs. fake motion latent  $z'$ . Finally at generation, we maps the learned latent $z'$ to motion  sequence using the decoder $\mathcal{D}(z')$ }
    \label{fig:2}
\end{figure}

Our method overview is shown in figure:\ref{fig:2}, where we adapt the conditional GAN \cite{mirza2014conditional} architecture to latent space. In particular, the generator $G$ takes the latent $z$ and conditioned input $c$ and generates the fake motion latent space $z' = G(z,c)$. On the other hand, discriminator $D$ learns to differentiate between real motion latent space $\mathcal{E}(x)$ and fake motion latent space $z'$ conditioned on $c$. We write the training objective of LS-GAN as a two-player min-max game with: 
$\min_{G} \max_{D} \mathbb{E}_{z \sim \varepsilon(x)} \left[ \log D(z, c) \right] + \mathbb{E}_{z \sim p_{z}(z)} \left[ \log(1 - D(G(z,c),c)) \right]$. During generation, we decode $z'$ into the motion sequences using the decoder $\mathcal{D}$, that is $\hat{x}^{1:L}$ = $\mathcal{D}(z')$ = $\mathcal{D}(G(z,c))$.  

\subsection{GAN architectures}
We experiment with two different LS-GAN architectures in the latent space setting. 

\textbf{Vanilla GAN: }
The Generator comprises three fully connected layers and the Discriminator consists of four fully connected layers.  Both models employ leaky ReLU activation to all layers preceding the final layer.

\begin{table*}[h]
\small
\vspace{2pt}
\begin{tabular}{@{}lccccccc@{}}
\toprule
{Methods} & \multicolumn{3}{c}{R Precision $\uparrow$}                                                                                                                & \multicolumn{1}{c}{{FID$\downarrow$}} & {MM Dist$\downarrow$}              & {Diversity$\rightarrow$}           & {MModality$\uparrow$}              \\ \cmidrule(lr){2-4}
              & \multicolumn{1}{c}{Top 1} & \multicolumn{1}{c}{Top 2} & \multicolumn{1}{c}{Top 3} & \multicolumn{1}{c}{}                     &                          &                            &                            \\ \midrule
Real &
  $0.511^{\pm.003}$ &
  $0.703^{\pm.003}$ &
  $0.797^{\pm.002}$ &
  $0.002^{\pm.000}$ &
  $2.974^{\pm.008}$ &
  $9.503^{\pm.065}$ &
  \multicolumn{1}{c}{-}
  \\ \midrule
Seq2Seq \cite{plappert2018learning} &
  $0.180^{\pm.002}$ &
  $0.300^{\pm.002}$ &
  $0.396^{\pm.002}$ &
  $11.75^{\pm.035}$ &
  $5.529^{\pm.007}$ &
  $6.223^{\pm.061}$ &
  \multicolumn{1}{c}{-} \\
LJ2P \cite{ahuja2019language2pose}&
  $0.246^{\pm.001}$ &
  $0.387^{\pm.002}$ &
  $0.486^{\pm.002}$ &
  $11.02^{\pm.046}$ &
  $5.296^{\pm.008}$ &
  $7.676^{\pm.058}$ &
  \multicolumn{1}{c}{-} \\
T2G\cite{bhattacharya2021text2gestures} &
  $0.165^{\pm.001}$ &
  $0.267^{\pm.002}$ &
  $0.345^{\pm.002}$ &
  $7.664^{\pm.030}$ &
  $6.030^{\pm.008}$ &
  $6.409^{\pm.071}$ &
  \multicolumn{1}{c}{-} \\
Hier \cite{ghosh2021synthesis}&
    $0.301^{\pm.002}$ &
    $0.425^{\pm.002}$ &
    $0.552^{\pm.004}$ &
    $6.532^{\pm.024}$ &
    $5.012^{\pm.018}$ &
    $8.332^{\pm.042}$ &
    \multicolumn{1}{c}{-} \\
TEMOS \cite{petrovich22temos}&
  $0.424^{\pm.002}$ &
  $0.612^{\pm.002}$ &
  $0.722^{\pm.002}$ &
  $3.734^{\pm.028}$ &
  $3.703^{\pm.008}$ &
  $8.973^{\pm.071}$ &
  $0.368^{\pm.018}$ \\
T2M \cite{Guo_2022_CVPR_t2m}&
  $0.457^{\pm.002}$ &
  $0.639^{\pm.003}$ &
  $0.740^{\pm.003}$ &
  $1.067^{\pm.002}$ &
  $3.340^{\pm.008}$ &
  $9.188^{\pm.002}$ &
  $2.090^{\pm.083}$ \\
MDM \cite{mdm2022human}&
  $0.320^{\pm.005}$ &
  $0.498^{\pm.004}$ &
  $0.611^{\pm.007}$ &
  $0.544^{\pm.044}$ &
  $5.566^{\pm.027}$ &
  $\boldsymbol{9.559}^{\pm.086}$ &
  $\underline{2.799}^{\pm.072}$ \\
 MotionDiffuse \cite{zhang2022motiondiffuse} &
  $\boldsymbol{0.491}^{\pm.001}$ &
  $\boldsymbol{0.681}^{\pm.001}$ &
  $\boldsymbol{0.782}^{\pm.001}$ &
  $0.630^{\pm.001}$ &
  $\boldsymbol{3.113}^{\pm.001}$ &
  $\underline{9.410}^{\pm.049}$ &
  $1.553^{\pm.042}$ \\
MLD \cite{chen2023executing} &
  $\underline{0.481}^{\pm.003}$ &
  $\underline{0.673}^{\pm.003}$ &
  $\underline{0.772}^{\pm.002}$ &
  $\boldsymbol{0.473}^{\pm.013}$ &
  $\underline{3.196}^{\pm.010}$ &
  $9.724^{\pm.082}$ &
  $2.413^{\pm.079}$  \\
 \midrule  

  Vanilla GAN  (Ours) &

 $0.327^{\pm.002}$ &
$0.492^{\pm.002}$ &
$0.599^{\pm.002}$ &
$1.507^{\pm.017}$ &
$3.994^{\pm.008}$ &
$9.320^{\pm.085}$ &
$0.313^{\pm.020}$



 \\
  Vanilla WGAN-GP (Ours) &

 $0.437^{\pm.002}$ &
$0.622^{\pm.002}$ &
$0.728^{\pm.002}$ &
$0.782^{\pm.016}$ &
$3.395^{\pm.007}$ &
$9.180^{\pm.085}$ &
$2.419^{\pm.091}$

 \\
 
  Deep GAN (Ours) &
 $0.352^{\pm.002}$ &
$0.531^{\pm.002}$ &
$0.645^{\pm.002}$ &
$3.036^{\pm.028}$ &
$3.907^{\pm.006}$ &
$8.631^{\pm.071}$ &
$0.308^{\pm.016}$
\\


Deep WGAN-GP (Ours) &
 $0.391^{\pm.002}$ &
$0.572^{\pm.002}$ &
$0.675^{\pm.002}$ &
$\underline{0.482}^{\pm.013}$ &
$3.731^ {\pm.014}$ &
$9.249^{\pm.067}$ &
$\boldsymbol{3.501}^{\pm.144}$



   \\ \bottomrule
\end{tabular}%
\vspace{-6pt}
\caption{Comparison of text-conditional motion synthesis on HumanML3D dataset. These metrics are evaluated by the motion encoder from \cite{Guo_2022_CVPR_t2m}. Empty MModality indicates the non-diverse generation methods. The right arrow $\rightarrow$ means the closer to real motion the better. \textbf{Bold} and \underline{underline} indicate the best and the second best result.} 
\label{tab:tm:comp:humanml3d}
\end{table*}

\textbf{Deep GAN: }
We add two residual blocks between the fully connected layers in both the Generator and discriminator architectures. Residual connections \cite{he2015deep} helps to train deeper networks by overcoming the vanishing gradients.



\begin{table*}[h] 
\centering
\begin{tabular}{@{}lrrrrlcccc@{}}
\toprule
\multirow{4}{*}{Methods} & \multicolumn{4}{c}{FLOPs (G) $\downarrow $} & \multirow{4}{*}{Parameter} & \multicolumn{4}{c}{FID $\downarrow $} \\
\cmidrule(lr){2-5} \cmidrule(lr){7-10}
&\multicolumn{3}{c}{DDIM}& DDPM&&\multicolumn{3}{c}{DDIM}&DDPM\\
\cmidrule(lr){2-4} \cmidrule(lr){5-5} \cmidrule(lr){7-9} \cmidrule(lr){10-10}
& 50 & 100 & 200 & 1000&&50&100 & 200&1000\\
\midrule
MDM
&597.97&1195.94&2391.89&11959.44&
$x \in \mathbb{R}^{196\times512}$
&7.334&5.990&5.936&0.544\\ 
MLD
&29.86&33.12&39.61&91.60& 
$z \in \mathbb{R}^{1\times256}$ 
&0.473&0.426&0.432&\textcolor{black}{0.568}\\
\midrule
Vanilla GAN
&\multicolumn{4}{c}{\textbf{1.581}}&
$z \in \mathbb{R}^{1\times100}$
&\multicolumn{4}{c}{0.783}\\
Deep GAN
&\multicolumn{4}{c}{\textbf{2.665}}&
$z \in \mathbb{R}^{1\times100}$
&\multicolumn{4}{c}{0.482}\\
\bottomrule
\end{tabular}
\caption{Evaluation of floating-point operations on text-to-motion.  We evaluate the FLOPs on 2048 motion clips, counted by THOP library.}
\label{tab:FLOPS}
\end{table*}

\section{Dataset, Loss and Evaluation metrics}
\subsection{Dataset}  

\textbf{Text-to-motion: }
\textbf{HumanML3D} \cite{guo2022generating}  is a 3D human motion-language dataset which covers a wide range of human actions including human activities like walking, jumping, swimming, playing golf etc. It contains 14,616 motion sequences from AMASS \cite{Mahmood2019AMASSAO} and annotates 44,970 sequence-level textual descriptions.  Here, we employ
the motion representation as combination of: 3D joint
rotations, positions, velocities, and foot contact.

\textbf{Action-to-motion: }
\textbf{HumanAct12} \cite{Guo2020Action2MotionCG} is a action-to-motion language dataset that provides 1,191 raw motion sequences and 12 action categories. 
 
    

\subsection{Loss }
We experiment with Binary Cross entropy (BCE) and Wasserstein loss \cite{arjovsky2017wasserstein}.  We use sigmoid activation on the discriminator with BCE loss.  We use Gradient penality \cite{gulrajani2017improved} instead of weight clipping in Wasserstein GAN.  

    


\subsection{Metrics}
To assess the performance of our models, we utilize metrics as in  MLD \cite{chen2023executing}: FID, R-precision,  Diversity, Multi-modality, Multimodal Distance (MM Dist), Average position error(APE), Average variance error (AVE). To measure the computational workload , we use FLOPs.  

\begin{table*}[h]
\centering
\begin{tabular}{@{}lcccc@{}}
\toprule
\multirow{2}{*}{Methods}                                             & \multicolumn{4}{c}{HumanAct12}                          \\ \cmidrule(lr){2-5} 
                     & {$\text{FID}_{\text{train}}\downarrow$} & ACC $\uparrow$& DIV$\rightarrow$ & MM$\rightarrow$ \\ 
\toprule

Real &
  $0.020^{\pm.010}$ &
  $0.997^{\pm.001}$ &
  $6.850^{\pm.050}$ &
  $2.450^{\pm.040}$ \\
  \midrule
ACTOR \cite{Petrovich2021ActionConditioned3H}&
  $0.120^{\pm.000}$ &
  $0.955^{\pm.008}$ &
  $\underline{6.840}^{\pm.030}$ &
  $\underline{2.530}^{\pm.020}$ \\
INR \cite{Cervantes2022ImplicitNR}&
  $\underline{0.088}^{\pm.004}$ &
  $\underline{0.973}^{\pm.001}$ &
  $6.881^{\pm.048}$ &
  $2.569^{\pm.040}$ \\
MDM \cite{Tevet2022HumanMD}&
  $0.100^{\pm.000}$ &
  $\boldsymbol{0.990}^{\pm.000}$ &
  $6.680^{\pm.050}$ &
  $\boldsymbol{2.520}^{\pm.010}$\\
MLD \cite{chen2023executing} &
$\boldsymbol{0.077}^{\pm.004}$&
$0.964^{\pm.002}$&
$6.831^{\pm.050}$&
$2.824^{\pm.038}$ \\
  \midrule
Deep WGAN-GP (Ours) &
$0.110^{\pm.004}$&
$0.942^{\pm.002}$&
$\boldsymbol{6.850}^{\pm.053}$&
$2.585^{\pm.057}$

\\ \bottomrule
\end{tabular}%
\vspace{-8pt}
\caption{Comparison of action-conditional motion synthesis on HumanAct12: $\text{FID}_{\text{train}}$ indicate the evaluated splits. Accuracy (ACC) for action recognition. Diversity (DIV), MModality (MM) for generated motion diversity within each action label. The right arrow $\rightarrow$ means the closer to real motion the better. \textbf{Bold} and \underline{underline} indicate the best and the second best result.}
\label{tab:comp:action}
\end{table*}

\section{Training details and Results}

\subsection{Implementation details }
We borrow the Motion transformer encoders $\mathcal{E}$ and
decoder $\mathcal{D}$ from in MLD \cite{chen2023executing}. Our VAE model consists of 9 layers and 4 heads with skip connections. To train VAE, we follow the same loss configuration as MLD. All our models are trained on A100 GPU with AdamW optimizer using a fixed learning rate of $10^{-4}$. Our batch size is set to 128
during the VAE training stage and 64 during the LS-GAN
training stage. We report the test metrics on the training checkpoints with the lowest FID.  In all of our experiments, we use the latent dimension $z \in \mathbb{R}^{1\times100}$, $z' \in \mathbb{R}^{1\times256}$. We choose $c \in \mathbb{R}^{1\times768}$ for text-to-motion task and $c \in \mathbb{R}^{1\times10}$ for action-to-motion task.

\subsection{Text-to-motion }
For text-to-motion, we utilize the VAE checkpoint from iteration 1250 and keep it fixed during GAN training. For detailed evaluation metrics of the VAE, refer table:\ref{tab:vae_comparison}. 
Figures:\ref{fig:1}, \ref{fig:7} show the qualitative results for the text-to-motion task with LS-GAN (Refer Appendix \ref{sec: app text} for more results). 
 Table:\ref{tab:tm:comp:humanml3d} summarizes the test metrics with mean and 95$\%$ confidence interval from
20 times running (most of the results are borrowed from MLD \cite{chen2023executing}). We observe that the vanilla and deep GAN architectures gave the best empirical metrics and qualitative results when used with wasserstein loss with gradient penality. Table:\ref{tab:FLOPS} depicts the total number of floating-point operations on 2048 motion clips.

Our Deep WGAN-GP achieves a FID of 0.482 that is near-parity with the state-of-the-art MLD \cite{chen2023executing} (FID of 0.473) with $91\%$ in FLOPs reduction as shown in  Table:\ref{tab:FLOPS}. It outperforms MDM \cite{mdm2022human} in R precision, FID, MM Dist, MModality. It also achieves state-of-the-art  across MModality compared to all the previous models. Furthermore, our LS-GAN outperforms cross-modal models such as Seq2Seq \cite{plappert2018learning}, 
LJ2P \cite{ahuja2019language2pose}, 
T2G\cite{bhattacharya2021text2gestures} , 
Hier \cite{ghosh2021synthesis},
TEMOS \cite{petrovich22temos}, T2M \cite{Guo_2022_CVPR_t2m} across all evaluation metrics. This signifies high-quality motion and high text prompt
matching while maintaining a rich motion diversity as evident in  Figures:\ref{fig:1}, \ref{fig:7}. These results demonstrate that a simple GAN in latent space can achieve impressive results with minimal compute in both training and inference compared to the Diffusion models.


\subsection{Action-to-motion }

The action-conditioned task involves generating motion sequences based on an input action label. We compare our Deep WGAN-GP with ACTOR \cite{Petrovich2021ActionConditioned3H}, INR \cite{Cervantes2022ImplicitNR}, MDM \cite{Tevet2022HumanMD}, and MLD \cite{chen2023executing}. ACTOR and INR are transformer-based VAE models specifically designed for the action-conditioned task. In contrast, MDM and MLD are diffusion models that utilize the same learnable action embedding module as our method. We report the test metrics as the mean and 95$\%$ confidence interval computed from 20 independent runs.

From table:\ref{tab:comp:action}, we observe that Deep WGAN-GP outperforms all the other models in Diversity while maintaining competitive performance on FID, accuracy and MultiModality (MM). These results indicate that GAN in motion latent can also benefit action-conditioned motion generation task.

\renewcommand{\arraystretch}{1.3} 
\begin{table}[!ht]
\centering
\small
\begin{tabular}{l|c|c}
\hline
Metric  & VAE 250 &  VAE 1250 \\
        & checkpoint & checkpoint \\
\hline


APE\_root/mean $\downarrow$ & $0.0897^{\pm0.0002}$ & $\boldsymbol{0.0756}^{\pm0.0002}$ \\
APE\_traj/mean $\downarrow$ & $0.0857^{\pm0.0002}$ & $\boldsymbol{0.0723}^{\pm0.0002}$ \\
APE\_mean\_pose/mean $\downarrow$ & $0.0379^{\pm0.0000}$ & $\boldsymbol{0.0312}^{\pm0.0000}$ \\
APE\_mean\_joints/mean $\downarrow$ & $0.1008^{\pm0.0002}$ & $\boldsymbol{0.0845}^{\pm0.0002}$ \\
AVE\_root/mean $\downarrow$ & $0.0221^{\pm0.0001}$ & $\boldsymbol{0.0201}^{\pm0.0001}$ \\
AVE\_traj/mean $\downarrow$& $0.0220^{\pm0.0001}$ & $\boldsymbol{0.0200}^{\pm0.0001}$ \\
AVE\_mean\_pose/mean $\downarrow$ & $0.0021^{\pm0.0000}$ & $\boldsymbol{0.0015}^{\pm0.0000}$ \\
AVE\_mean\_joints/mean $\downarrow$ & $0.0241^{\pm0.0001}$ & $\boldsymbol{0.0216}^{\pm0.0001}$ \\
R\_precision\_top\_1 $\uparrow$ & $0.4422^{\pm0.0030}$ & $\boldsymbol{0.4891}^{\pm0.0020}$ \\
R\_precision\_top\_2 $\uparrow$ & $0.6337^{\pm0.0020}$ & $\boldsymbol{0.6803}^{\pm0.0023}$ \\
R\_precision\_top\_3 $\uparrow$ & $0.7379^{\pm0.0025}$ & $\boldsymbol{0.7787}^{\pm0.0021}$ \\
FID $\downarrow$ & $1.1754^{\pm0.0030}$ & $\boldsymbol{0.2661}^{\pm0.0010}$ \\
Diversity $\rightarrow$ & $\boldsymbol{9.3856}^{\pm0.0843}$ & $9.6901^{\pm0.0990}$ \\
MultiModality $\uparrow$ & $\boldsymbol{0.2056}^{\pm0.0095}$ & $0.1237^{\pm0.0058}$ \\

\hline
\end{tabular}
\caption{Comparison of evaluation metrics (mean and 95$\%$ confidence interval from running 20 times) on text-to-motion VAE 250, 
 $1250^{\text{th}}$ checkpoint. 
\textbf{Bold} indicate the best result.}
\label{tab:vae_comparison}
\end{table}

\begin{figure*}[h]
\centering
\resizebox{\linewidth}{!}{%
    \begin{tabular}{c}
 \subfloat[\Huge Visualization of the t-SNE results on evolved latent codes during the reverse diffusion process on action-to-motion task.
    $t$ is the diffusion step but ordered in the forward diffusion trajectory.
     ]{\includegraphics[width= 80cm]{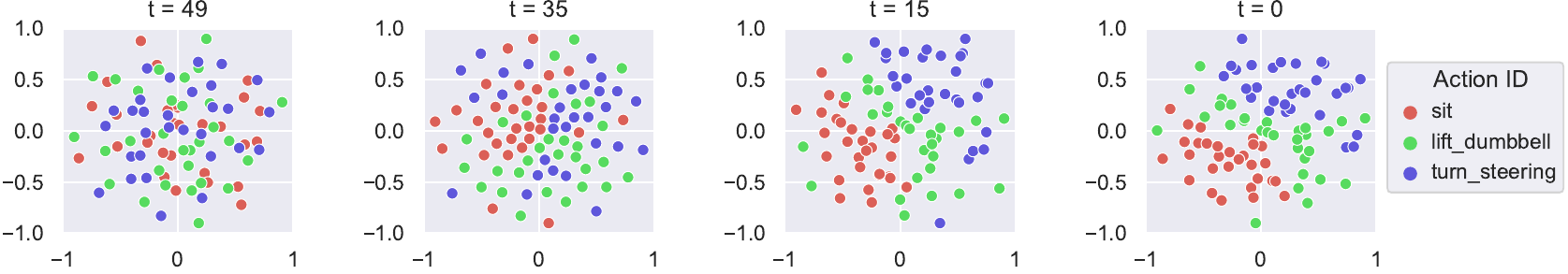} } \\ 
    
    \subfloat[\Huge Visualization of the t-SNE results on latent codes for Vanilla GAN, Vanilla WGAN-GP, Deep GAN, Deep WGAN-GP  (left to right) ]{\includegraphics[width=80cm, trim= 40 350 40 100 ,clip ]{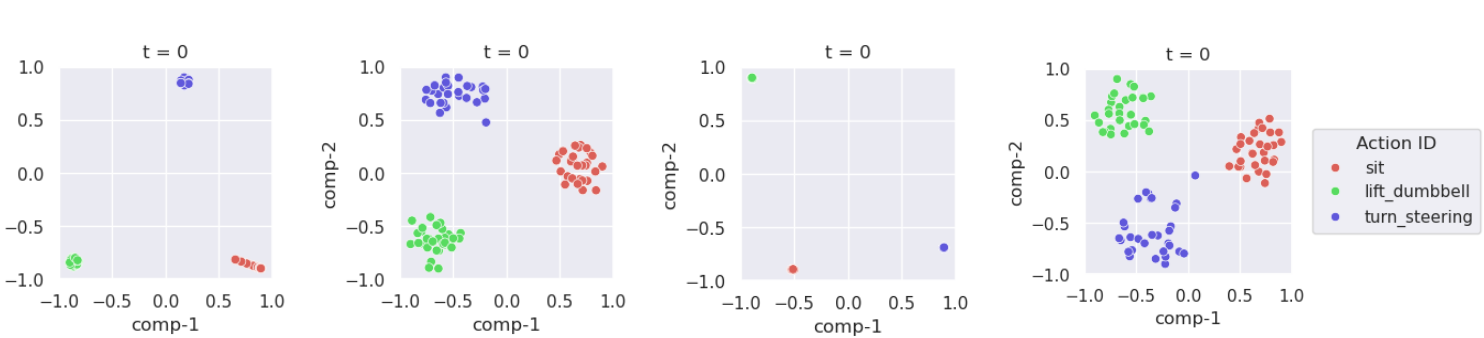} }
    \end{tabular}}
    \caption{Visualization of the t-SNE results on latent codes  of LS-GAN compared to MLD. We sample 30 motions for each action label}
    \label{fig:tsnse}
\end{figure*}

\section{Latent space visualization}
In this section, we present t-SNE visualizations of the latent space on action-to-motion task, illustrating how our LS-GAN effectively captures and separates different actions within the latent space. These results are compared with the MLD \cite{chen2023executing} in Figure \ref{fig:tsnse}.  

From the latent space visualization, it is evident that Vanilla GAN and Deep GAN have low MultiModality scores (measures the generation diversity within the same text or action input), while Vanilla WGAN-GP and Deep WGAN-GP have higher MultiModality highlighting the effectiveness of the Wasserstein loss with gradient penalty. This observation even holds true for the text-to-motion generation task, as evidenced by the results presented in Table \ref{tab:tm:comp:humanml3d}. Furthermore, our approach shows superior separation of latent code clusters at timestep $t=0$ compared to MLD. This improved clustering at $t=0$ indicates that our LS-GAN framework captures a more structured  motion latent representation, potentially leading to better interpretability and generation fidelity.


\section{Discussion} \label{sec:6}
\subsection{Addressing GAN challenges:}   Usage of condition information in our model helps us to overcome  mode collapse challenge by conditioning the model on additional information. In addition, our generator learns the inherent features of real motion data. This encourages the discriminator to compare the underlying properties instead of the high dimensional real data, similar to Feature Matching \cite{salimans2016improvedtechniquestraininggans} that helps to stabilize training. Methods such as regularization, spectral normalization , adaptive learning rates, multiple generators/discriminators, auxiliary loss as discussed by \cite{saxena2023generativeadversarialnetworksgans} can be further explored to stabilize GAN training. 

\subsection{Accelerated Diffusion: }
 Diffusion distillation \cite{yin2024onestepdiffusiondistributionmatching, sauer2023adversarialdiffusiondistillation, luo2023comprehensivesurveyknowledgedistillation} is a knowledge distillation task, where
a student model is trained to distill the multi-step outputs
of the original diffusion model into a single or few steps. These prior works, require a separate pretraining and distillation phase. In addition, one-step diffusion models require a greater attention in choosing the training objectives and scheduling mechanism \cite{yin2024onestepdiffusiondistributionmatching}.  On the other hand, recent works on GANs \cite{sauer2023stylegantunlockingpowergans, kang2023scalingganstexttoimagesynthesis} shows that StyleGAN-T, Giga-Gan outperforms Distilled diffusion models on text-to-image generation. Considering these, we believe GAN in latent space would serve as a solution to accelerated diffusion.

\subsection{Limitations}
First, similar to most motion generation methods, our approach can generate motion sequences of arbitrary lengths, but still below the maximum length in the dataset. Secondly, LS-GAN specifically targets human body motion, in contrast to works focusing on facial motion \cite{Karras2017AudiodrivenFA} or hand motion \cite{Li2021PIANOAP}. Lastly, we limited ourselves to simple prompts for text-to-motion. It may be beneficial to consider the impact of motion outputs on edge cases and ambiguous text descriptions.


\section{Conclusion}
In this paper, we introduced a novel approach for text-to-motion and action-to-motion synthesis using Generative Adversarial Networks in the latent space. By leveraging the power of GANs and the compact representation of motion sequences in the latent space, our method achieves faster training and inference times compared to previous methods while maintaining high-quality motion synthesis results. Results demonstrate that a simple GAN in latent space is comparable to complex models. This work will open a new direction in exploring latent space GANs that can have faster stable training and inference compared to latent space diffusion.

\clearpage

{\small
\bibliographystyle{ieee_fullname}
\bibliography{egbib}
}

\clearpage 

\section{Appendix}

\subsection{LS-GAN qualitative results on text-to-motion: }  \label{sec: app text}


\begin{figure}[!ht]
\centering
\resizebox{\textwidth}{!}{%
    \begin{tabular}{ccc}
        \subfloat[\LARGE A person is skipping rope.  ]{{\includegraphics[width=10cm]{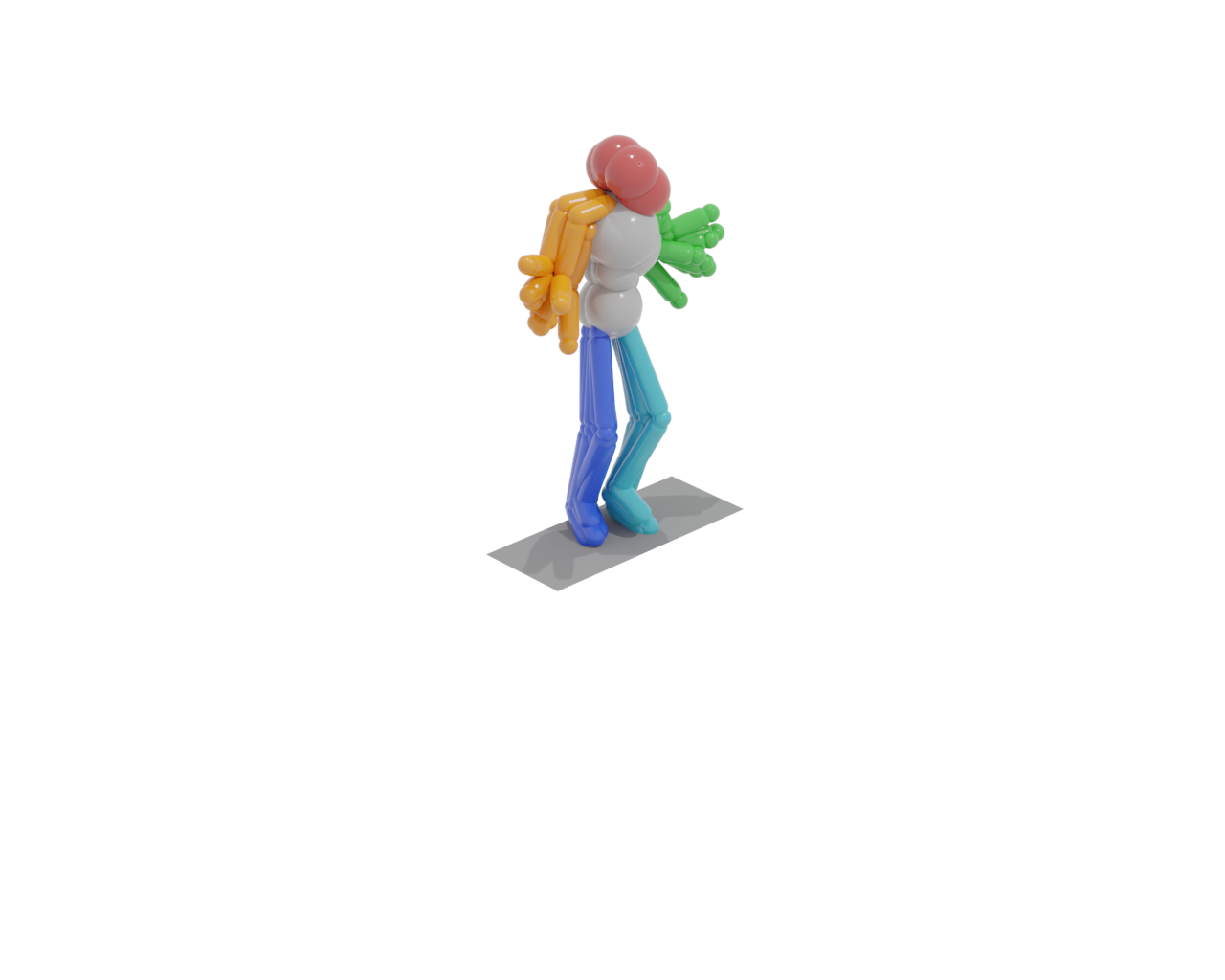} }} 
        \subfloat[\LARGE a person doing jumping jacks.   ]{{\includegraphics[width=10cm]{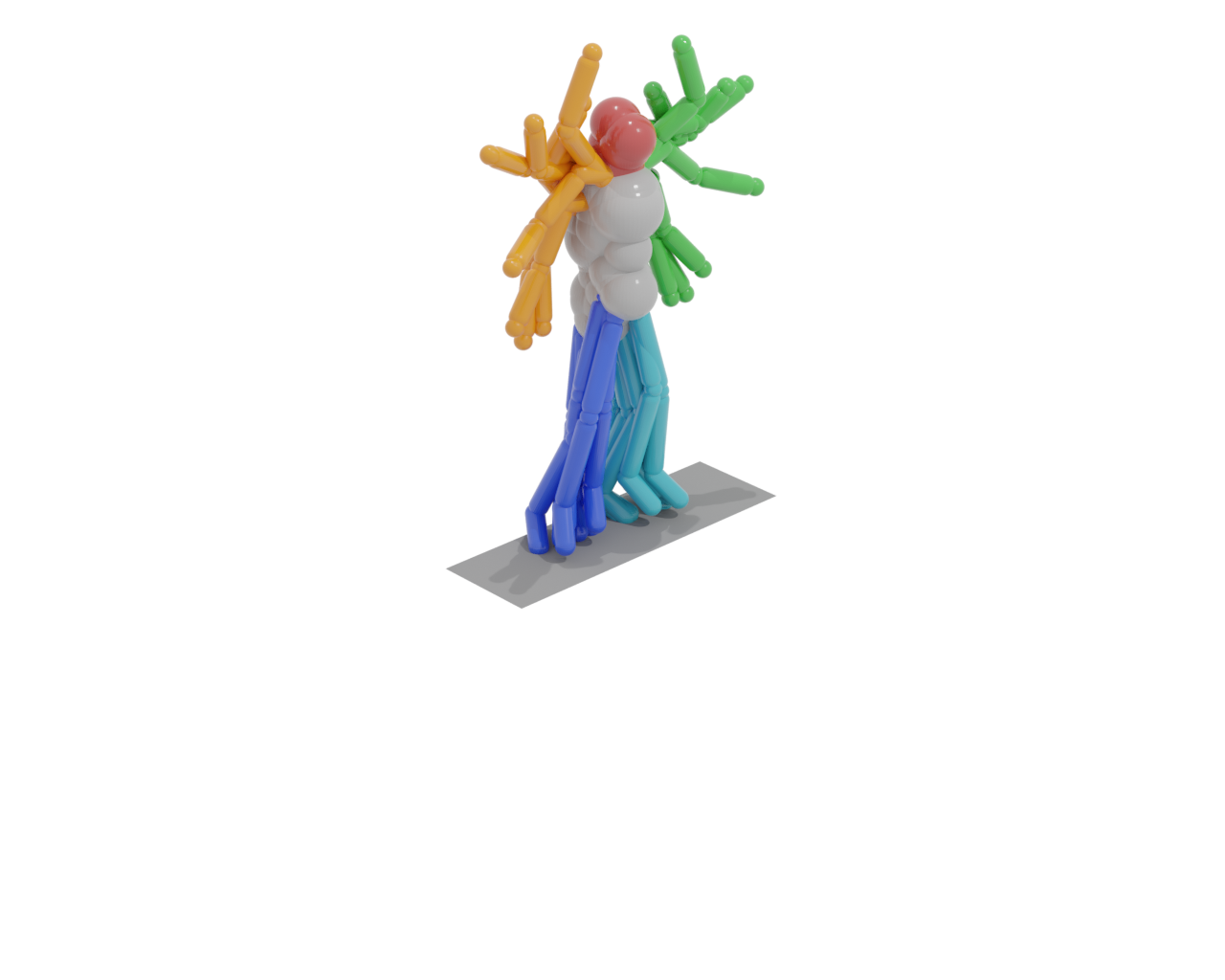} }} 
        \subfloat[\LARGE a person jogs straight forward. ]{{\includegraphics[width=10cm]{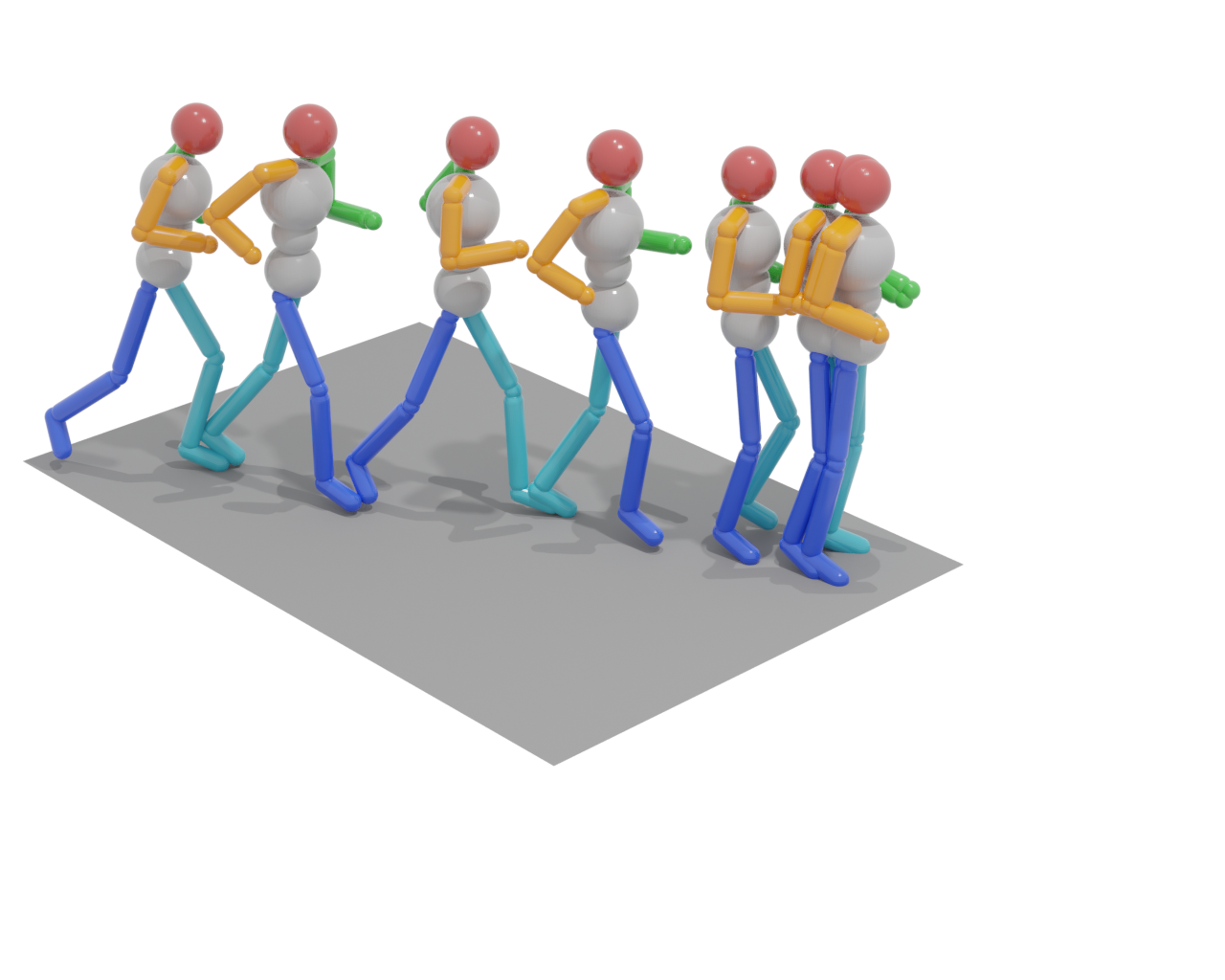} }} 
    \end{tabular}}
    \parbox{\textwidth}{\caption{Qualitative results of our method shown by Vanilla GAN}}
    \label{fig:3}
\end{figure}

\begin{figure}[!ht]
\centering
\resizebox{\textwidth}{!}{%
    \begin{tabular}{ccc}
        \subfloat[\LARGE a man kicks with something or someone with his left leg. ]{{\includegraphics[width=10cm]{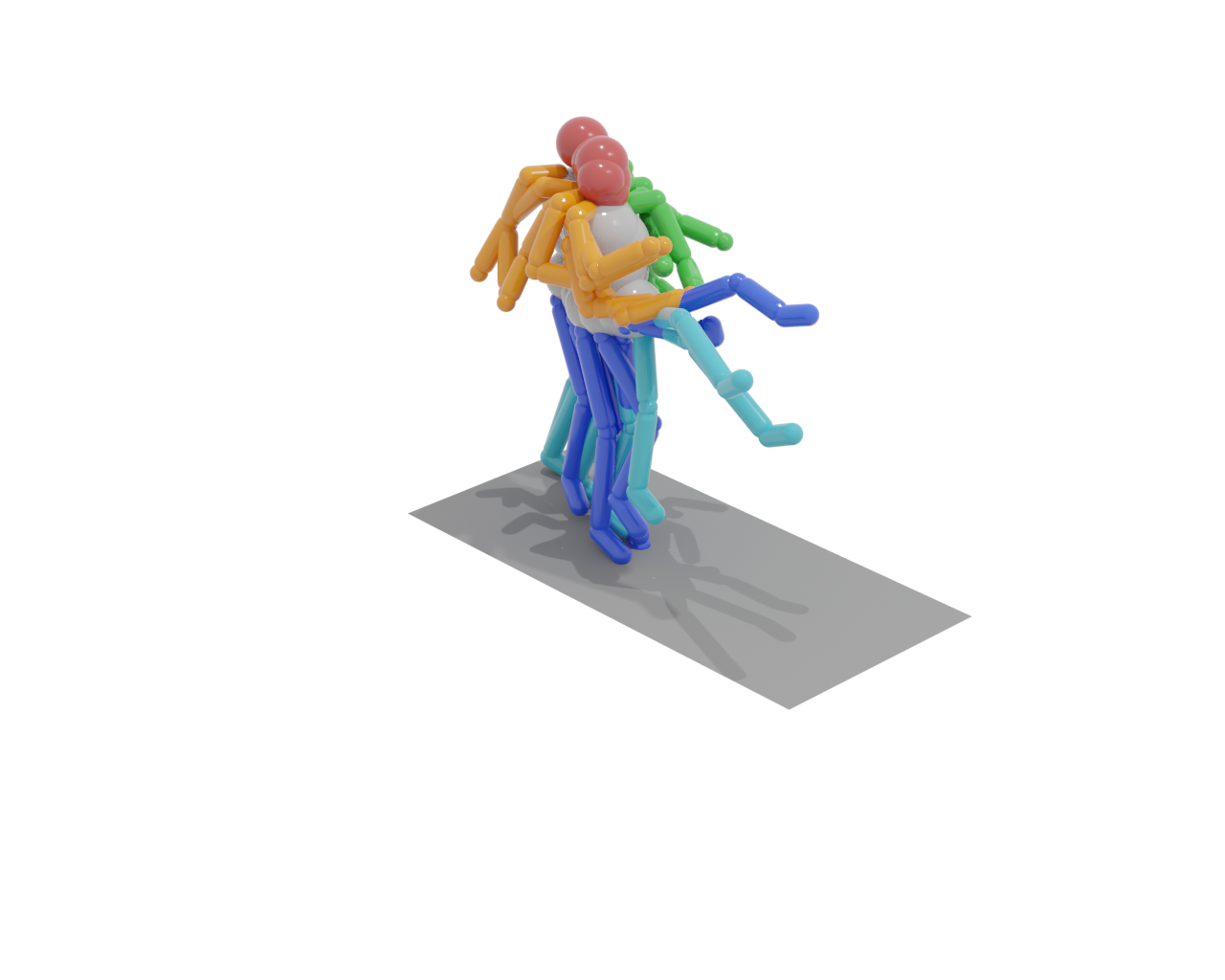} }} 
        \subfloat[\LARGE a person walks backward slowly. ]{{\includegraphics[width=10cm]{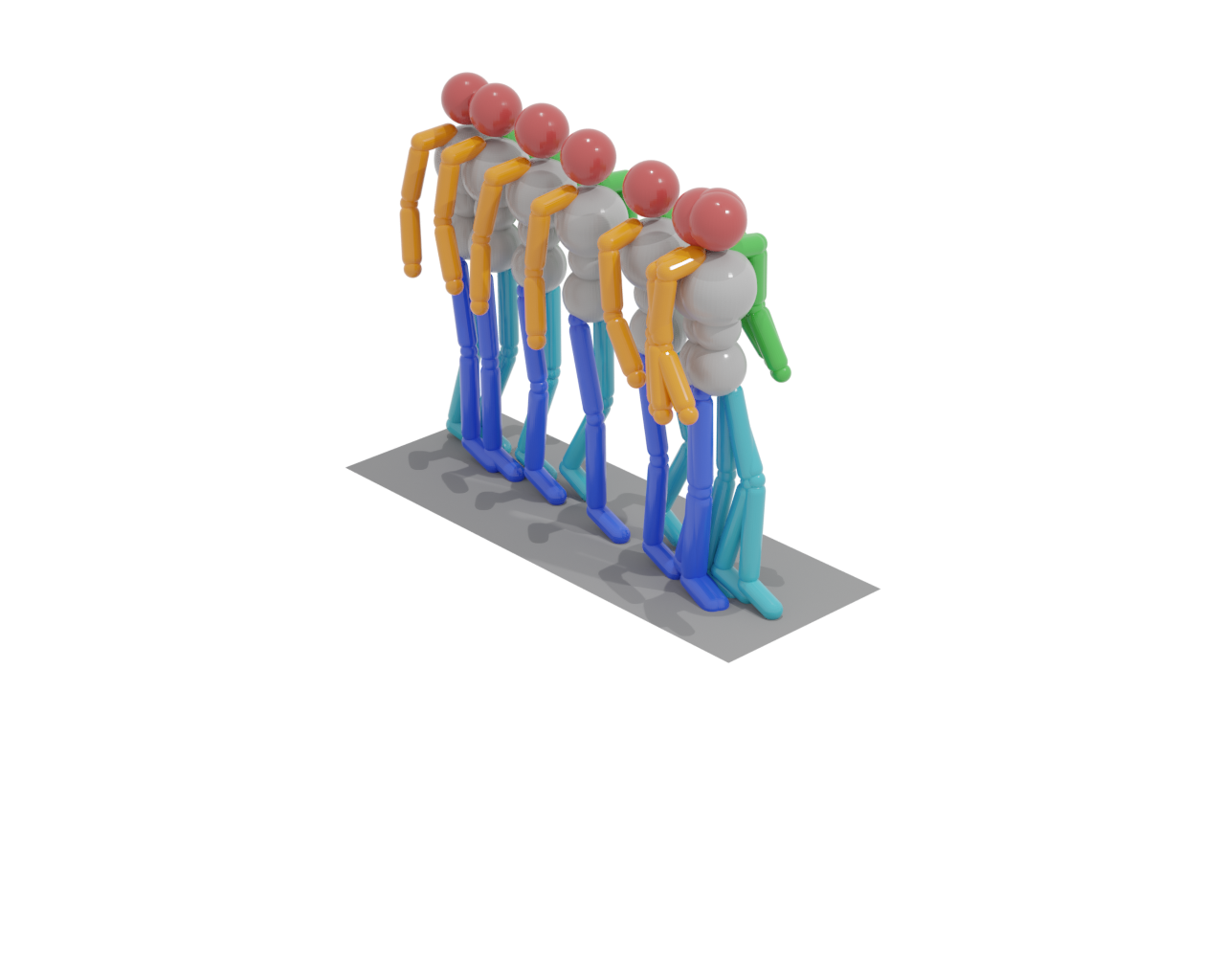} }} 
        \subfloat[\LARGE a person jogs straight forward. ]{{\includegraphics[width=10cm]{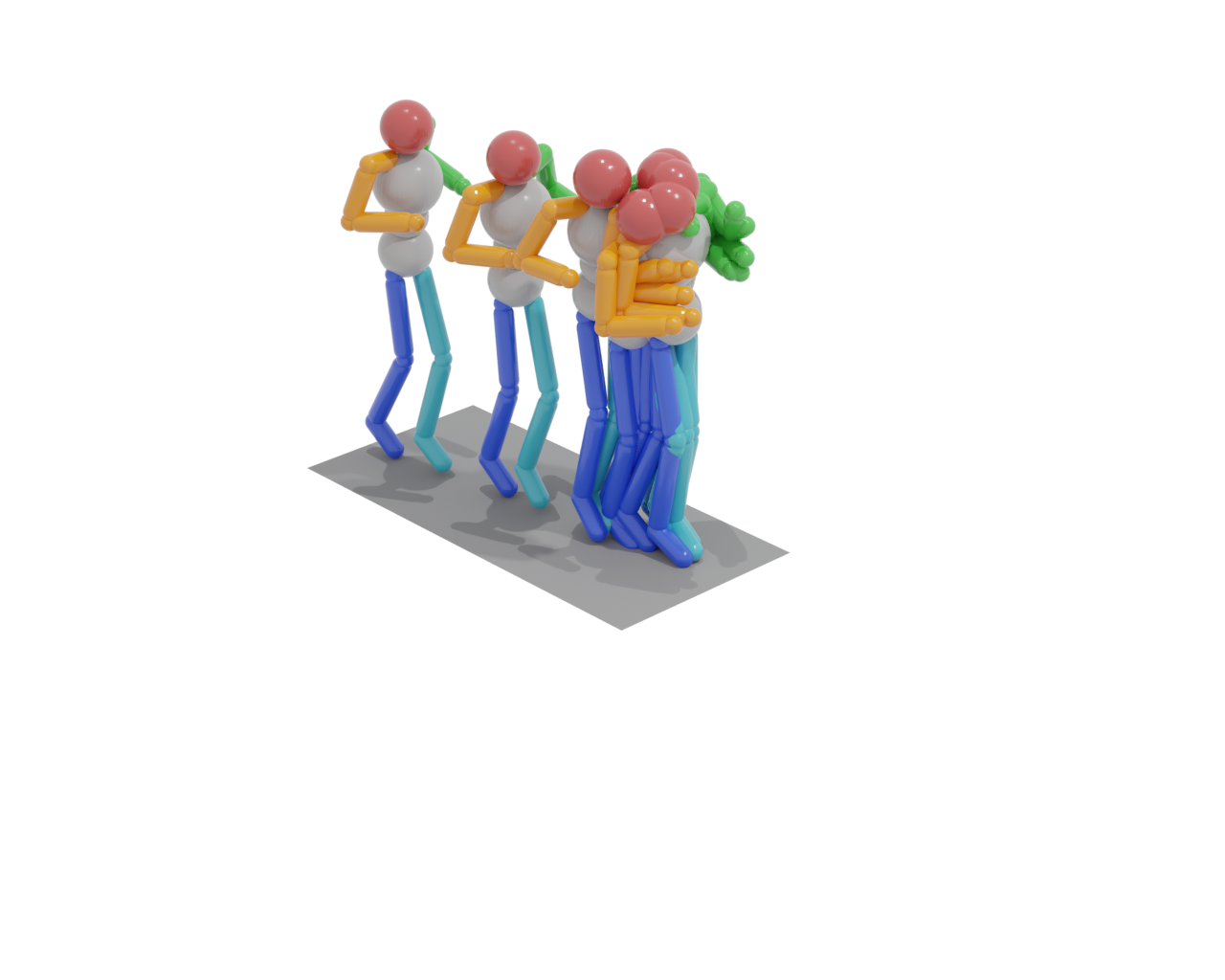} }} 
    \end{tabular}}
    \parbox{\textwidth}{\caption{Qualitative results of our method shown by Deep GAN}}
    \label{fig:4}
\end{figure}

\begin{figure}[!ht]
\centering
\resizebox{\textwidth}{!}{%
    \begin{tabular}{ccc}
        \subfloat[\LARGE a man kicks with something or someone with his left leg.  ]{{\includegraphics[width=10cm]{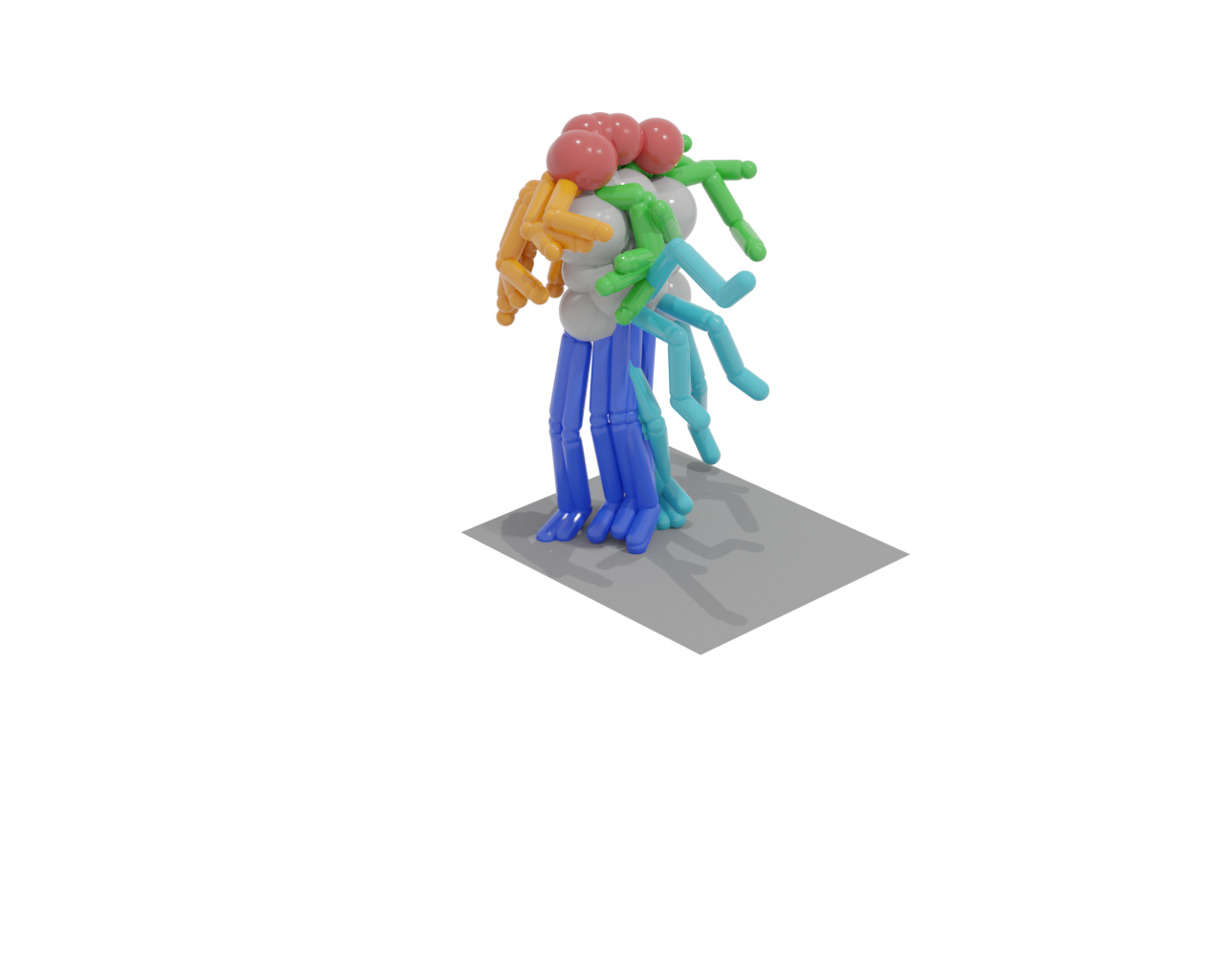} }} 
        \subfloat[\LARGE a person doing jumping jacks.  ]{{\includegraphics[width=10cm]{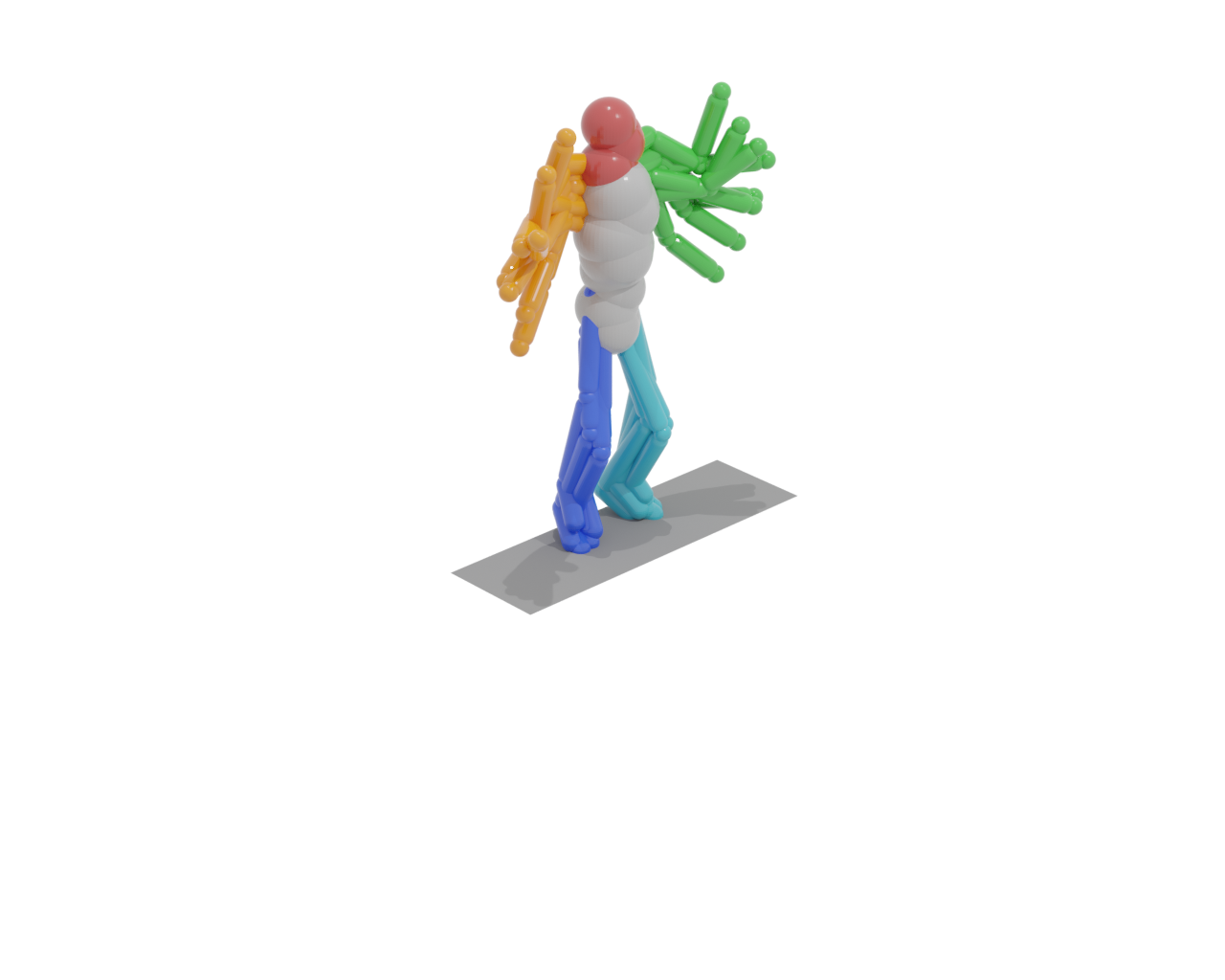} }} 
        \subfloat[\LARGE a person jogs straight forward.  ]{{\includegraphics[width=10cm]{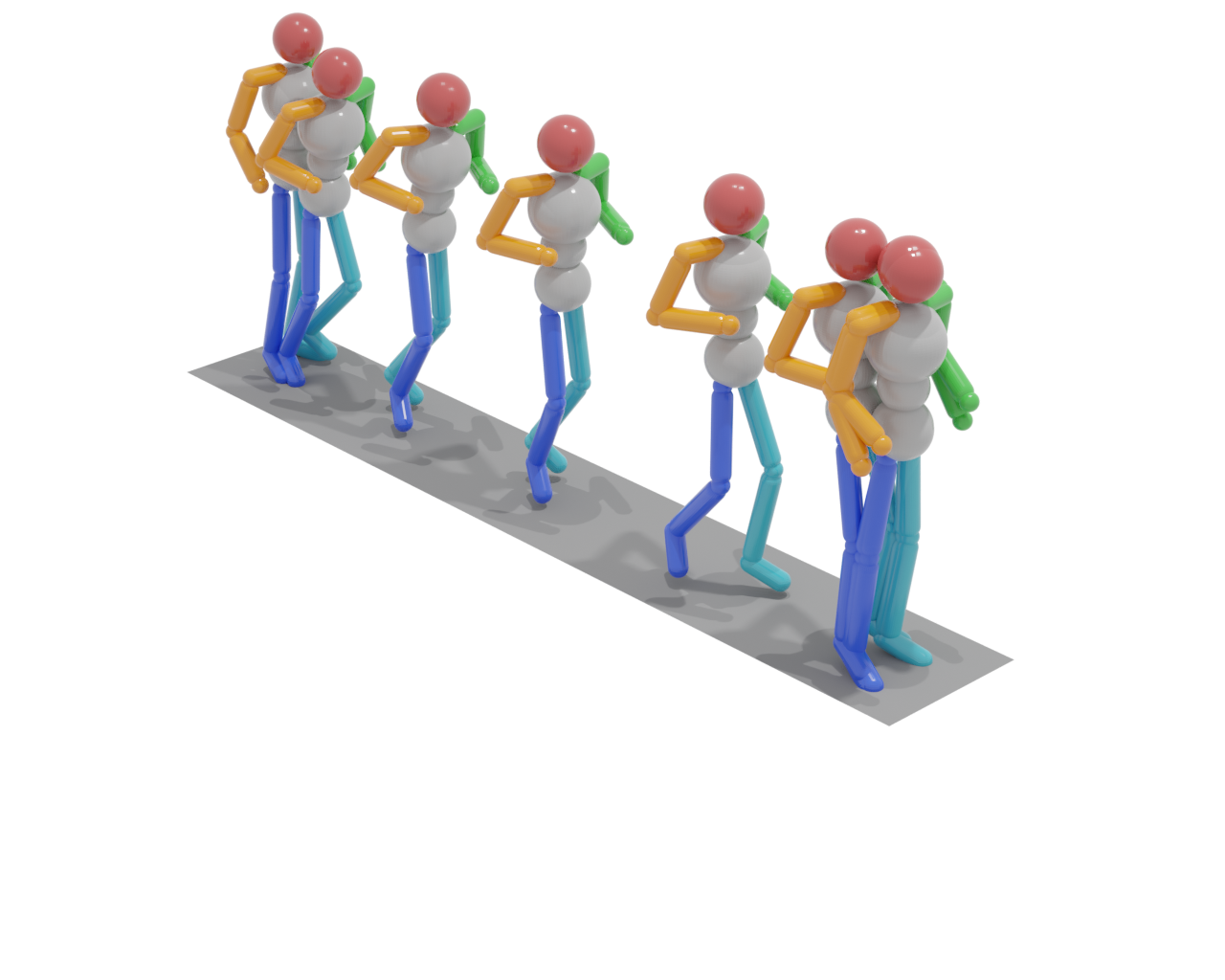} }} 
    \end{tabular}}
    \parbox{\textwidth}{\caption{Qualitative results of our method shown by Vanilla WGAN-GP}}
    
    \label{fig:5}
\end{figure}

\clearpage

\begin{figure}[!ht]
\centering
\resizebox{\textwidth}{!}{%
    \begin{tabular}{ccc}
        \subfloat[\LARGE a person walks backward slowly. ]{{\includegraphics[width=10cm]{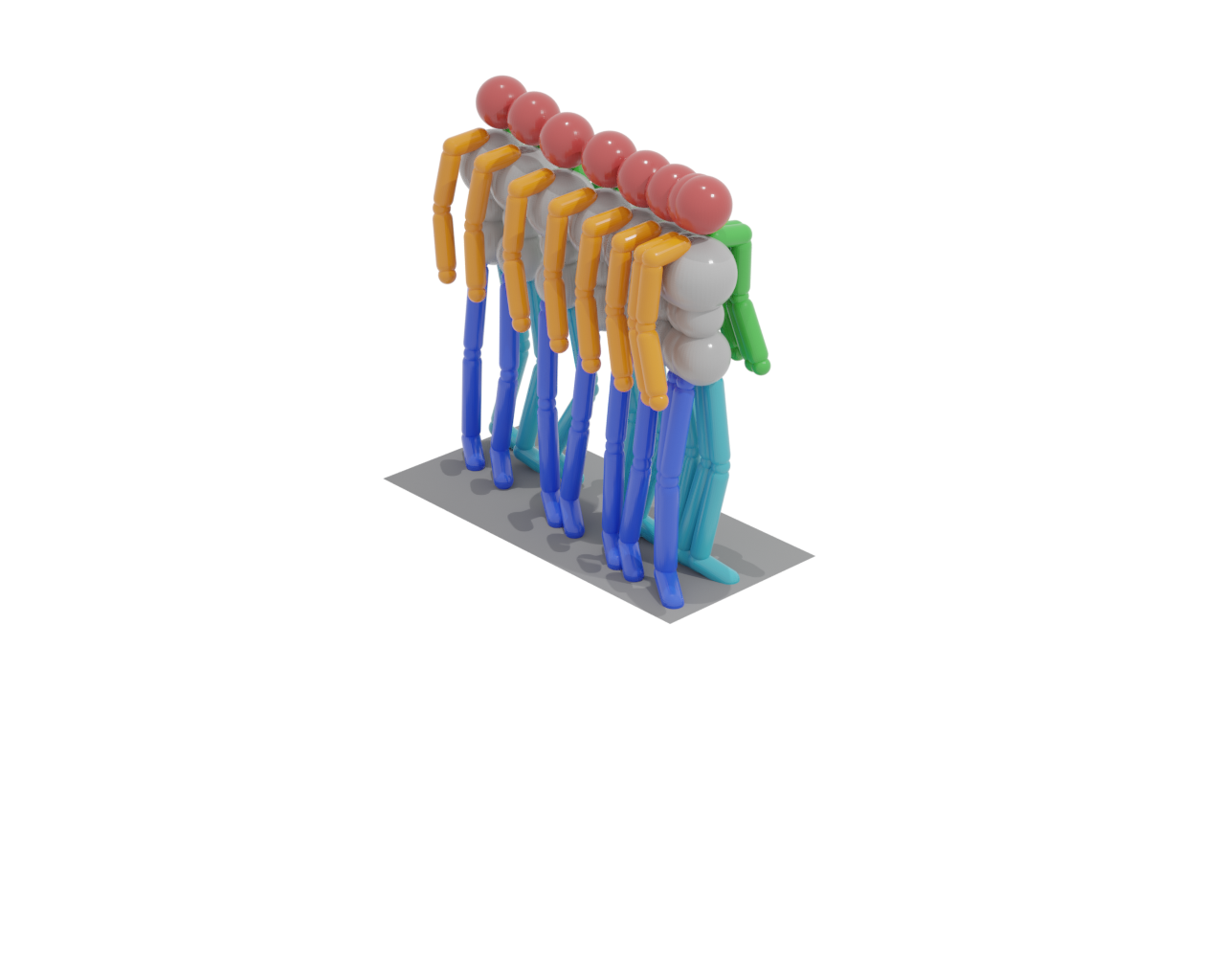} }} 
        \subfloat[\LARGE a person raised arms up and pull them down. ]{{\includegraphics[width=10cm]{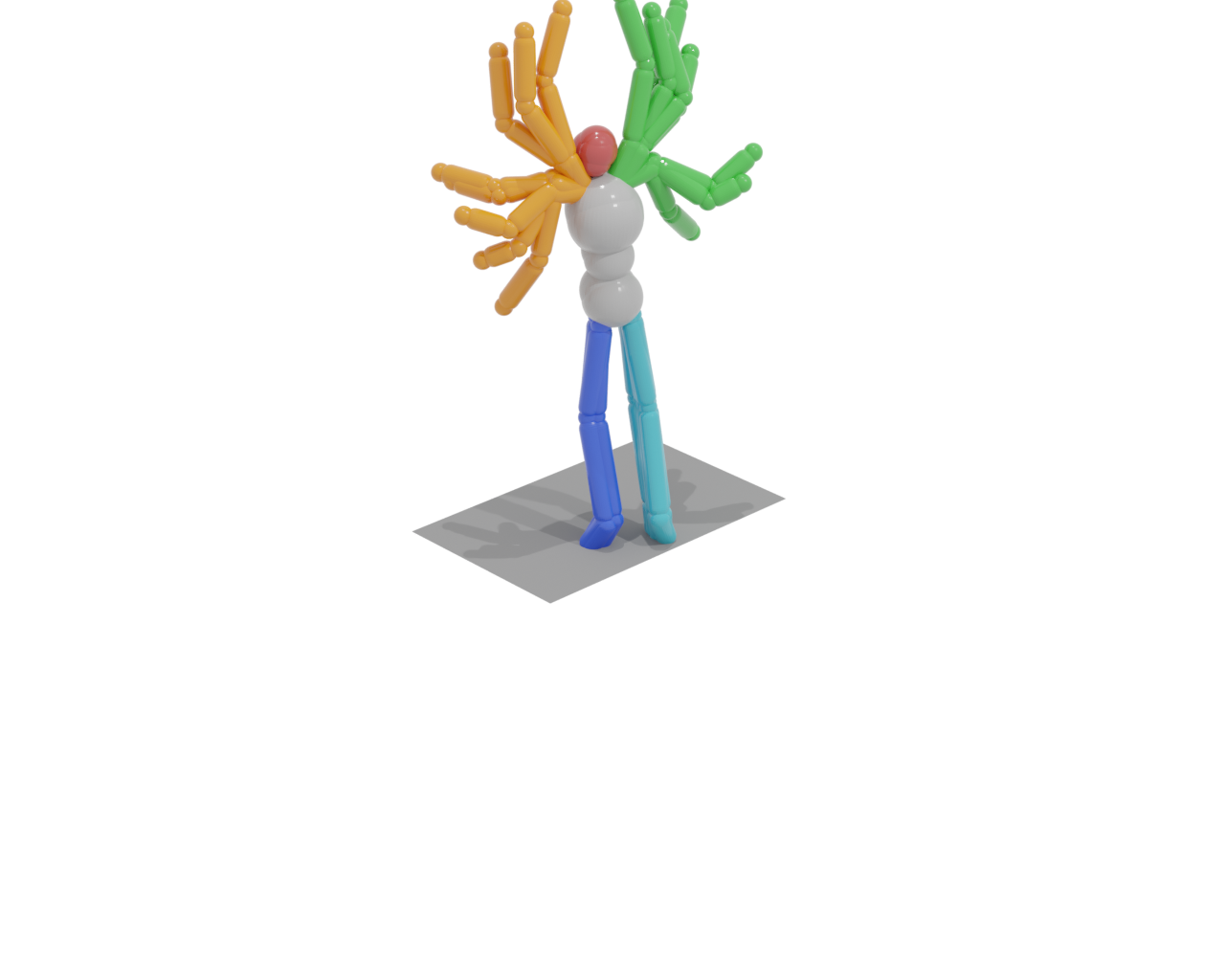} }} 
        \subfloat[\LARGE a person walking forward with legs wide apart.]{{\includegraphics[width=10cm]{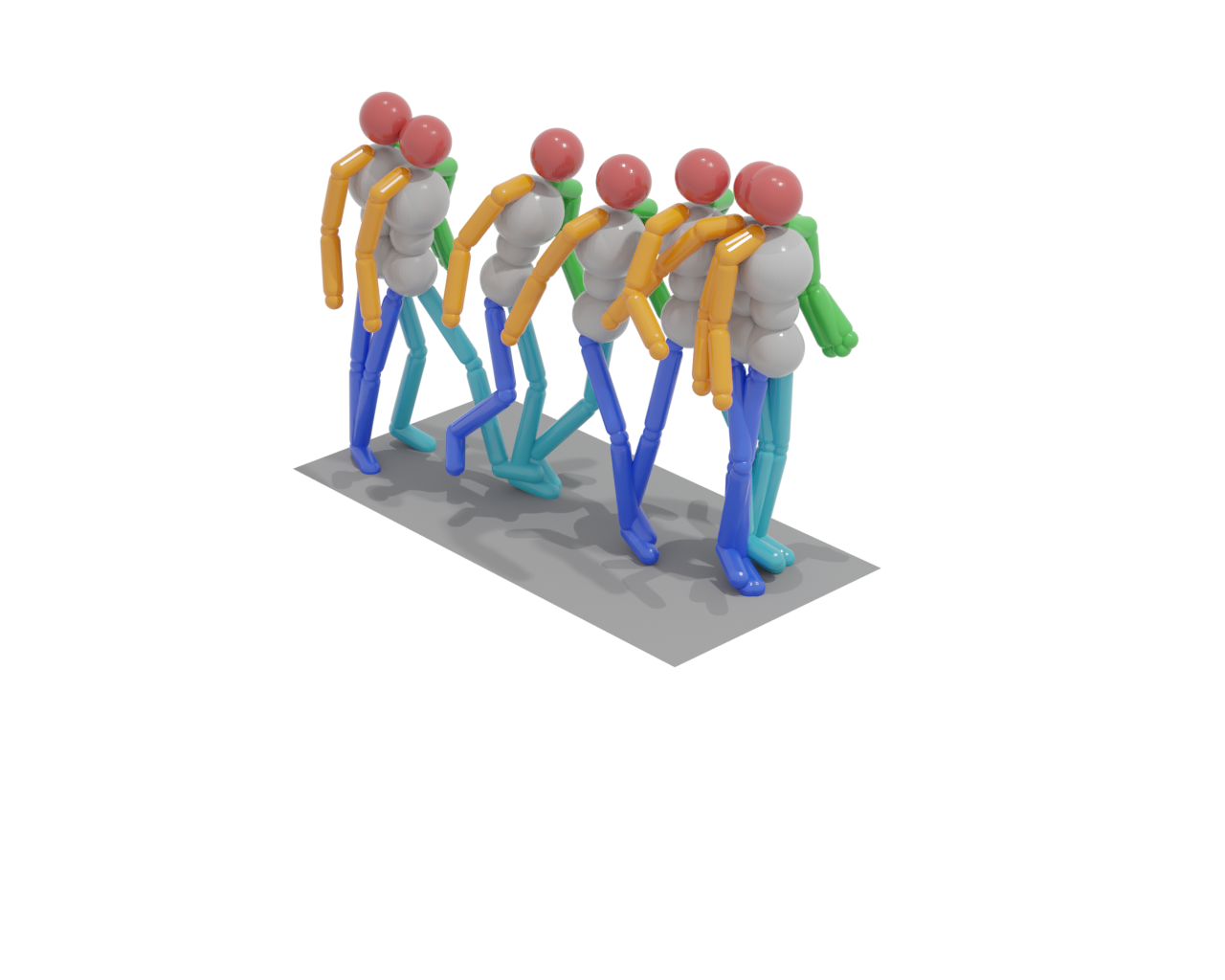} }} 
    \end{tabular}}
    \parbox{\textwidth}{\caption{Qualitative results of our method shown by Deep WGAN GP}}
    \label{fig:6}
\end{figure}


       

\end{document}